\documentclass{article}

    \PassOptionsToPackage{numbers, compress}{natbib}

\usepackage{wrapfig}
\usepackage{PRIMEarxiv}
\usepackage[utf8]{inputenc} 
\usepackage[T1]{fontenc}    
\usepackage{hyperref}       
\usepackage{url}            
\usepackage{booktabs}       
\usepackage{amsfonts}       
\usepackage{amsthm}

\usepackage{nicefrac}       
\usepackage{microtype}      
\usepackage{xcolor}         
\usepackage{natbib} 
    \renewcommand{\bibsection}{\subsubsection*{References}}
\usepackage{graphicx}
\usepackage{amsmath,amssymb}
\usepackage{amsfonts} 
\usepackage[utf8]{inputenc}       
\usepackage{amsmath, amssymb, amsfonts}
\usepackage{bm}                   
\usepackage{csquotes}             
\usepackage{amsthm,amsmath,amsfonts,amssymb}

\usepackage{enumitem}

\usepackage{bm}   

\usepackage[ruled,vlined,linesnumbered]{algorithm2e}
\SetKwComment{Comment}{\color{gray}// }{}
\DontPrintSemicolon            

\usepackage{xspace}         
\newcommand{\CTAE}{\textsc{CTAE}\xspace}

\title{
 Coupled Transformer Autoencoder for Disentangling Multi-Region Neural Latent Dynamics
}

\author{
  Ram Dyuthi Sristi \\
  UC San Diego \\
  \texttt{rsristi@ucsd.edu}
\And
  Sowmya Manojna Narasimha \\
  UC San Diego \\
  \texttt{smnarasimha@ucsd.edu}
\And
  Jingya Huang \\
  UC San Diego \\
  \texttt{jih201.ucsd@gmail.com}
\AND
  Alice Despatin \\
  RWTH Aachen University $\&$ FZ Jülich\\
  \texttt{alice.despatin@rwth-aachen.de} \\
\And
  Simon Musall \\
  RWTH Aachen University  $\&$ FZ Jülich\\
  \texttt{s.musall@fz-juelich.de} \\
\AND
  Vikash Gilja \\
  UC San Diego \\
  \texttt{vgilja@ucsd.edu}
\And
  Gal Mishne \\
  UC San Diego \\
  \texttt{gmishne@ucsd.edu}
}

\begin{document}

\maketitle

\begin{abstract}

Simultaneous recordings from thousands of neurons across multiple brain areas reveal rich mixtures of activity that are shared between regions and dynamics that are unique to each region. Existing alignment or multi-view methods neglect temporal structure, whereas dynamical latent-variable models capture temporal dependencies but are usually restricted to a single area, assume linear read-outs, or conflate shared and private signals.  
We introduce Coupled Transformer Autoencoder (\CTAE{})—a sequence model that addresses both (i) non-stationary, non-linear dynamics and (ii) separation of shared versus region-specific structure,
in a single framework. 
\CTAE{} employs Transformer encoders and decoders to capture long-range neural dynamics, and explicitly partitions each region’s latent space into orthogonal shared and private subspaces. 
We demonstrate the effectiveness of \CTAE{} on two high-density electrophysiology datasets of simultaneous recordings from multiple regions, one from motor cortical areas and the other from sensory areas. \CTAE{} extracts meaningful representations that better decode behavior variables compared to existing approaches.

\end{abstract}

\section{Introduction}\label{sec:intro}

The advent of high-density electrophysiology probes, e.g., Neuropixels, and volumetric calcium imaging has enabled recording large-scale, high-resolution, multi-region neuronal datasets. 
This shift from single-area recordings to \emph{distributed circuits} reveals both globally coordinated signals and region-specific specialization \citep{jun2017neuropixels,machado2022multiregion}.  
Individual neurons both receive input from and project to multiple distant areas; behaviorally relevant codes are hypothesized to be broadcast across widespread circuits rather than confined to a canonical locus \citep{machado2022multiregion}.
This perspective challenges conclusions drawn from single-area studies and raises the question of which computations are local, which are distributed, and how global brain states are coordinated.   
Disentangling the \emph{shared} components that mediate inter-area interactions from \emph{private} signals unique to each region is therefore essential—both for mechanistic insight and for designing causal experiments such as targeted optogenetic inactivation or electrical stimulation of an upstream circuit.

Thus, recent work has focused on analyzing \emph{distributed circuits} to recover latent activity patterns both within and across regions.  
A successful multi-region latent model faces three simultaneous challenges: (i) latent trajectories must evolve smoothly in time to respect neural autocorrelation; (ii) they must accommodate the non-stationary, nonlinear dynamics that real circuits display; and (iii) they must separate shared from region-specific structure without a parameter explosion as the number of areas grows.

\begin{figure}[t]
    \centering
    \includegraphics[width=0.85\linewidth]{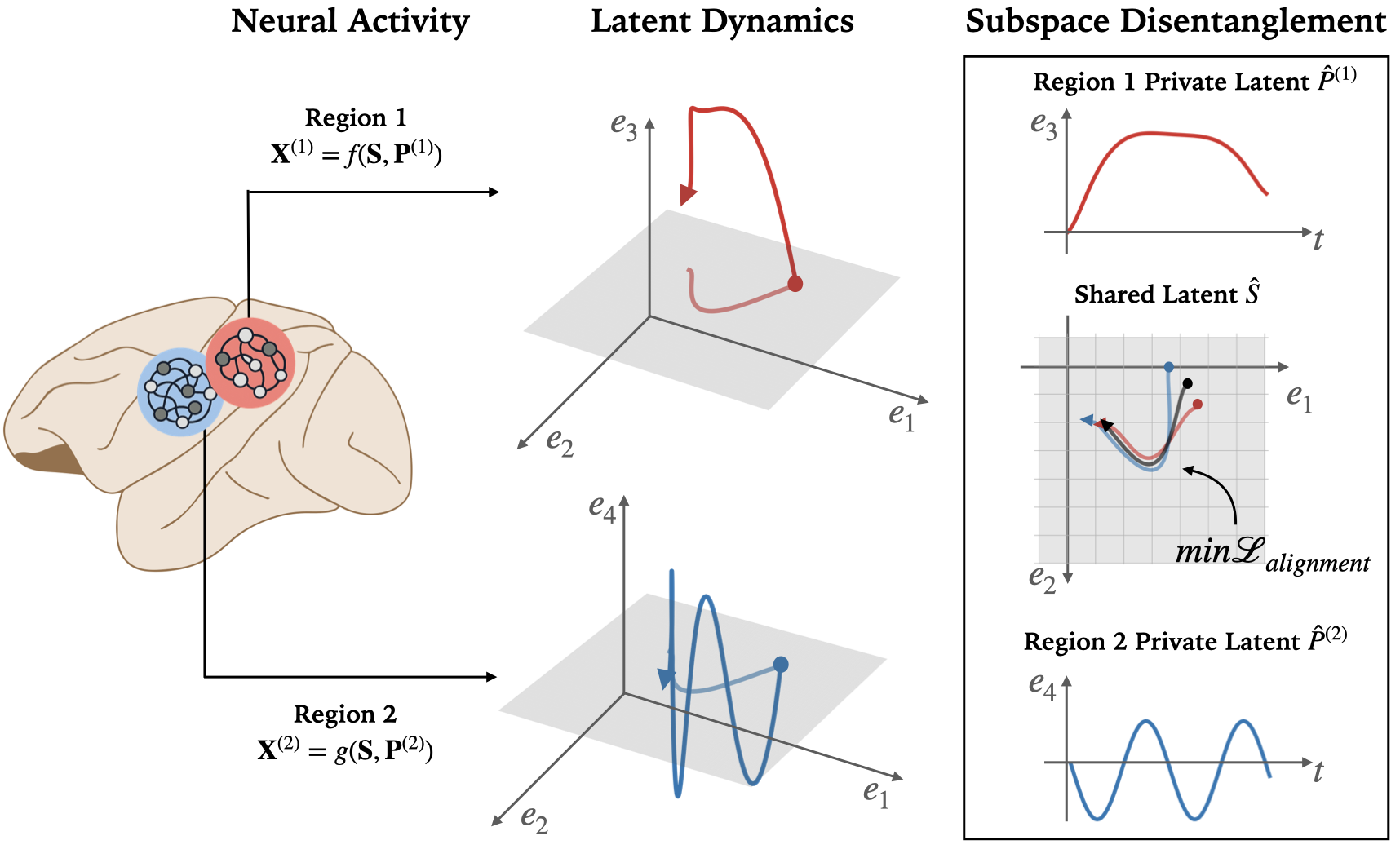}
    \caption{Observed neural activity across time from two brain regions, denoted as $\bm{X}^{(1)}$ and $\bm{X}^{(2)}$, is modeled as a nonlinear function of underlying latent dynamics specific to each region. In the illustration, $[e_1, e_2, e_3]$ span the latent subspace for region 1 and $[e_1, e_2, e_4]$ for region 2. Inter-regional communication is mediated by shared latent trajectories $S$ within the common subspace $[e_1, e_2]$, which drive correlated population activity. Following the output-null/potent hypothesis, we assume shared and private dimensions are orthogonal, allowing clear recovery of shared dynamics from region-specific processes.}
    \label{fig:overview}
\end{figure}

Even within a single brain region, recordings of neural population activity reveal a mixture of structured responses and substantial trial-to-trial variability, introducing a challenge when interrogating the behavior of neural circuits  \citep{cunningham2014dimred}. A widely adopted framework—neural latent dynamics \citep{vyas2020computation, churchland2024preparatory} —proposes that these high-dimensional responses reflect the evolution of a consistent, low-dimensional trajectory, supported by stable patterns of co-variability across neurons.   
Classical PCA or factor analysis (FA) uncover low-dimensional structure\citep{cunningham2014dimred} but ignore time. Moreover, because observed spikes are subject to high Poisson noise, the resulting latents can exhibit high-frequency dynamics that are difficult to interpret and may not reflect meaningful underlying processes. Consequently, latent variable models that incorporate temporal smoothness or dynamical constraints have been developed, based on either linear Gaussian Process (GP) models~\citep{yu2009gpfa} or nonlinear deep learning models~\citep{pandarinath2018lfads,ye2021ndt}.
Yet, when these single-area tools are applied naïvely to multi-area data, e.g., by concatenating the recordings, they often fail. Inter-area delays warp the latent space, differences in correlation structure cause shared factors to absorb private variance, and the larger or more active region can dominate the mixture weights. To address these challenges, recent studies have investigated both predictive models \citep{zandvakili2015coordinated, semedo2019communication, perich2018neural} and joint latent variable \citep{gokcen2022dlag} approaches for inter-regional activity. These efforts reveal that only a selective subset of latent dimensions actively participates in inter-area communication. Such findings motivate extensions to the latent dynamics framework, proposing that communication is mediated through a persistent, low-dimensional subspace—referred to as a communication subspace \citep{semedo2019communication}—that is distinct and orthogonal to private, region-specific dynamics Fig. \ref{fig:overview}.

More recently, joint latent variable models inspired by canonical correlation analysis (CCA) have been developed to capture correlated latent dynamics across regions while simultaneously learning region-specific latent components. One approach focused on neural data has been \emph{linear} approaches that generalize GP-based models to the multi-region setting~\citep{
gokcen2022dlag,gokcen2023mdlag,gokcen2024fast, li2024multi}. Alternatively, nonlinear autoencoder-based disentangling methods~\citep{lee2021dmvae, koukuntla2024splice} have been developed to analyze multiview point cloud data and partition latents into shared and private factors.
However, these treat time points as independent and identially distributed samples and therefore discard dynamics.  

\vspace{2pt}
\noindent\textbf{This work.}  
We introduce the \emph{Coupled Transformer Autoencoder} (\CTAE{}), an end-to-end framework for modeling simultaneous recordings from multiple brain areas.  
Transformer-based~\citep{vaswani2017attention} encoders and decoders act as flexible priors capable of capturing non-stationary, long-range neural dynamics; each region’s latent space is split into \emph{orthogonal shared and private subspaces}, with reconstruction losses preserving region-specific information and a lightweight alignment loss matching the shared representations.  
Because the latents are behaviour-agnostic, downstream decoding\textemdash whether kinematics, forces, or cognitive variables\textemdash can be carried out on the same embedding without retraining.
Our contributions are as follows:
\begin{itemize}[leftmargin=1.5em,itemsep=2pt,topsep=1pt]
\item \textbf{Non-stationary multi-region modelling.} \CTAE{} extracts shared and private representations that capture long-range, nonlinear dynamics absent in previous approaches. 
\item \textbf{Scalable architecture.} A mixing weight allows the shared subspace to extend trivially beyond two regions without an exponential increase in parameters.  
\item \textbf{Generic downstream utility.} The behaviour-agnostic latent space supports diverse decoding tasks (e.g.\ position, velocity, cognitive state) using simple linear read-outs.  
\item \textbf{Empirical validation.} On neural recordings from M1–PMd (Utah Arrays) and SC-ALM (Neuropixel Probes), \CTAE{} achieves higher shared-variance capture than existing works, while revealing interactions consistent with anatomy.
\end{itemize}

\section{Related work}\label{sec:related}

\noindent\textbf{Single-region latent-variable models.}  
Linear techniques—including PCA~\citep{pearson1901liii} and FA~\citep{harman1976modern} are widely used to analyze large-scale recordings~\citep{cunningham2014dimred}. 
Gaussian-Process Factor Analysis (GPFA) adds a smooth GP prior to FA to enforce temporal continuity \citep{yu2009gpfa}.  
GPFA’s stationarity and linear read-out assumptions are relaxed by Latent Factor Analysis via Dynamical Systems (LFADS) \citep{pandarinath2018lfads} and the Neural Data Transformer (NDT) \citep{ye2021ndt}, which exploit nonlinear recurrent generators or self-attention to capture non-stationary and nonlinear dynamics. Subsequent efforts pursued identifiability with switching LDS~\citep{johnson2016slds,glaser2022recurrent}, GP-SLDS~\citep{xu2024gpslds}, locally linear manifold models in DFINE~\citep{pernice2024dfine}, and variational neural state-space models such as VIND~\citep{hernandez2018nonlinear} and S4-based sequence layers~\citep{gu2022efficiently}.  

\vspace{4pt}
\noindent\textbf{Multi-region latent models.} 
Multi-region methods relying on GP models  include Delayed Latents Across Groups (DLAG)~\citep{gokcen2022dlag}, its multi-population generalization mDLAG~\citep{gokcen2023mdlag}, multi-view GPFA extensions~\citep{gokcen2024fast}, and 
Multi-Region Markovian Gaussian Process~\citep{ li2024multi}.
These inherit smooth-GP assumptions and linear read-outs that struggle with non-stationary or long-range dependencies. 
Classical multiset CCA~\citep{hotelling1936cca} and tensor‐decomposition methods~\citep{cichocki2015tensor} have also been applied to multi-area recordings, however these are linear techniques. 
Current-Based Decomposition (CURBD)~\citep{perich2020curbd} infers directed information flow between many regions from data-constrained RNNs. Unlike the above methods, CURBD does not explicitly model a shared–private decomposition of latent dynamics.

\vspace{4pt}
\noindent\textbf{Multiview autoencoders}  
Multi-view representation learning has progressed beyond classical CCA by incorporating additional losses and constraints to more effectively identify shared latent representations. However, these methods often require adaptation to account for the specific statistical and temporal properties of neural data.  
Correlation-based alignments—canonical correlation analysis (CCA) and its deep variant (DCCA) \citep{andrew2013dcca}—maximise instantaneous correlation but provide no guarantee of capturing \emph{all} shared variance; Deep CCA Autoencoders (DCCAE) add reconstructions at the risk of mixing private with shared information \citep{wang2016dccae}.
Representative examples include Deep Coupled Auto-encoder Networks~\citep{wang2014deeply}, Coupled Autoencoders for domain adaptation~\citep{wang2023generalized}, Correlated Autoencoders for audiovisual retrieval~\citep{feng2014cross}, Deep Correlation Autoencoders (DCCAE)~\citep{wang2016dccae}, Split-brain Autoencoders~\citep{zhang2017split}, shared-private domain adaptation AEs~\citep{bousmalis2016domain} and cycle-consistent multiview AEs \citep{wu2018cycle}.  SPLICE \citep{koukuntla2024splice} extends this line with measurement-network penalties to sharpen disentanglement. However, it scales poorly—its auxiliary \emph{measurement networks} grow exponentially with region count. On the other hand, DMVAE assumes one global shared component across \emph{all} views, precluding subset-specific latents. 
Multimodal VAE (MVAE)~\citep{wu2018multimodal}, Disentangled Multimodal VAE (DMVAE) \citep{lee2021dmvae}, Joint Multimodal VAE \citep{sutter2021generalized}, 
Mixture-of-Experts MVAE \citep{shi2019variational} learn shared/private factors across images, text and audio but treat each sample independently, ignoring sequential dependencies.

\noindent Across these categories, no existing method jointly satisfies \emph{(i)} non-stationary nonlinear dynamics, \emph{(ii)} temporal continuity and \emph{(iii)} scalability to more than two regions without incurring a parameter explosion as the number of areas increases.


\section{Problem formulation}
\label{sec:problem}

Let $\bm{X}^{(1)} \in \mathbb{R}^{N_{1} \times T}, \bm{X}^{(2)} \in \mathbb{R}^{N_{2} \times T}$ denote simultaneous neural population recordings from two brain regions, each acquired over $T$ time steps and, $N_1$ and $N_2$ channels respectively (generalization to more than 2 regions is in Appendix~\ref{sec:ctae_multi}).
We assume that the observed neural activity at each time step $t$ is a nonlinear transformation of latent variables (Fig.\ref{fig:model}), which include: (i) shared latent dynamics within a communication subspace \citep{semedo2019communication} spanned by neural trajectories that are correlated across regions, and (ii) private latent dynamics that capture region-specific processes lying in subspaces orthogonal to the shared component. To effectively capture the temporal structure of neural activity, we model these latent variables as functions of the entire observed neural activity.
\begin{equation}
    \mathbf{X}_t^{(1)} = f\!\bigl(\bm{S}_{1:t}, \bm{P}^{(1)}_{1:t}\bigr), \quad \quad 
    \mathbf{X}_t^{(2)} = g\!\bigl(\bm{S}_{1:t}, \bm{P}^{(2)}_{1:t}\bigr)
  \qquad \text{for } t \in \{1,\ldots,T\}
\end{equation}
where ${1:t}$ denotes the sequence of indices between 1 and $t$, $\bm{S}_t \in \mathbb{R}^{d_s}$ represents the
\emph{shared} dynamics between the two regions at time $t$, and
$\bm{P}^{(1)}_t \in \mathbb{R}^{d_1}$ and
$\bm{P}^{(2)}_t \in \mathbb{R}^{d_2}$ capture dynamics
specific to regions 1 and 2, respectively, and
$d_s$, $d_1$, and $d_2$ denote the dimensionalities of the
shared and private representations.  

Our goal is to recover the shared and private latent dynamics $\bm{S}$, $\bm{P}^{(1)}$, and
$\bm{P}^{(2)}$ given the observed neural activity
$\mathbf{X}^{(1)}$ and $\mathbf{X}^{(2)}$.

\noindent \textbf{Notations.}
Bold upper‑case letters denote matrices
(e.g., $\mathbf{X}$), bold lower‑case denote latent vectors (e.g.,  $\bm{w}$),
and $\widehat{\cdot}$ indicates reconstructions.
Let $\mathbf{1}_{n}$ denote a vector of all-ones of length $n$ and let $\mathbf{0}_{n}$ denote a vector of all zeros of length $n$. $\Vert (\cdot) \Vert_F$ denotes the Frobenius norm. 
$\odot$ denotes element-wise multiplication.
Expectation is over batches of trials unless stated otherwise.

\section{Coupled Transformer Autoencoders}
\label{sec:two_region}

Our model is comprised of a coupled autoencoder built on transformer models, that disentangles shared and private representations. 
We have designed loss functions to recover latents that maximally represent the neural activity, where the shared representations are aligned across all regions, and each disentangled sub-space is orthogonal to the others. For clarity, we describe the two-region case here; the extension to an arbitrary number of regions is briefly explained in Sec.~\ref{sec:r3} with full details provided in Appendix~\ref{sec:ctae_multi}.

\subsection{Model Architecture}

\label{subsec:arch_two}

We design separate causal Transformer-based encoder–decoder pairs, denoted by $(E_\theta^{(1)}, D_\phi^{(1)})$ and $(E_\theta^{(2)}, D_\phi^{(2)})$, for each of the two brain regions. Each encoder is a Transformer stack with
\emph{self-attention} layers that capture long-range, nonlinear temporal
dependencies within a region. Each decoder employs standard Transformer
\emph{cross-attention} to reconstruct the original firing rates from its
region’s latents. Processing the full multichannel time series from
each session, the encoders produce latent representations partitioned into two distinct components: 
shared latent sequences capturing dynamics common to both regions and private latent sequences specific to each region:
\begin{equation}
\mathbf{Z}^{(1)} = E_\theta^{(1)}\!\bigl(\mathbf{X}^{(1)}\bigr), \quad\quad 
\mathbf{Z}^{(2)} = E_\theta^{(2)}\!\bigl(\mathbf{X}^{(2)}\bigr), 
\quad\quad 
\mathbf{Z}^{(1)},\mathbf{Z}^{(2)} \in \mathbb{R}^{D \times T} 
\end{equation}
Here $D$ is the total number of latent dimensions. To indicate which of the latent dimensions correspond to shared and private representations, we introduce weight vectors $\mathbf{w}_1$ and $\mathbf{w}_2 \in \{0,1\}^{D}$.

\noindent \textbf{Region–specific weight masks.}
Let the latent dimension be partitioned as
\(D = d_s + d_1 + d_2\).
Define three contiguous index sets
\(\mathcal{I}_s = \{1,\ldots,d_s\} \),
\(\mathcal{I}_1 = \{d_s{+}1,\ldots,d_s{+}d_1\}\),
and
\(\mathcal{I}_2 = \{d_s{+}d_1{+}1,\ldots,D\} \).
We then construct binary weight vectors that indicate which subset of the latent dimensions correspond to the shared or to region-private. Define  
\(\mathbf{w}_1, \mathbf{w}_2 \in \{0,1\}^{D}\) as
\begin{equation}
\mathbf{w}_1
  =\bigl[\underbrace{\mathbf{1}_{d_s}}_{\mathcal{I}_s},
          \underbrace{\mathbf{1}_{d_1}}_{\mathcal{I}_1},
          \underbrace{\mathbf{0}_{d_2}}_{\mathcal{I}_2}\bigr]^\top,
\qquad
\mathbf{w}_2
  =\bigl[\underbrace{\mathbf{1}_{d_s}}_{\mathcal{I}_s},
          \underbrace{\mathbf{0}_{d_1}}_{\mathcal{I}_1},
          \underbrace{\mathbf{1}_{d_2}}_{\mathcal{I}_2}\bigr]^\top.
\label{eq:w_def}
\end{equation}
so that $\mathbf{w}_1$ activates the shared and region-private of region 1
dimensions, whereas $\mathbf{w}_2$ activates the shared and
region-private dimensions of region 2.
Throughout the paper we treat $\mathbf{w}_r$ as fixed;
the latent space dimensions $(d_s,d_1,d_2)$ are treated as
hyper-parameters and tuned on a validation set. Importantly, these masks do not hard-code the actual interaction structure:
they only specify an \emph{upper bound} on how many shared or private latents
the model may allocate. During training, dimensions unsupported by the data
naturally collapse to negligible variance. While fixed masks provide clarity
and control in the two-region setting, they could also be initialized from
anatomical priors or made learnable so that the model itself infers which
latents are shared versus private. We leave these scalable extensions to
future work.

\begin{figure}
    \centering
    \includegraphics[width=0.9\linewidth]{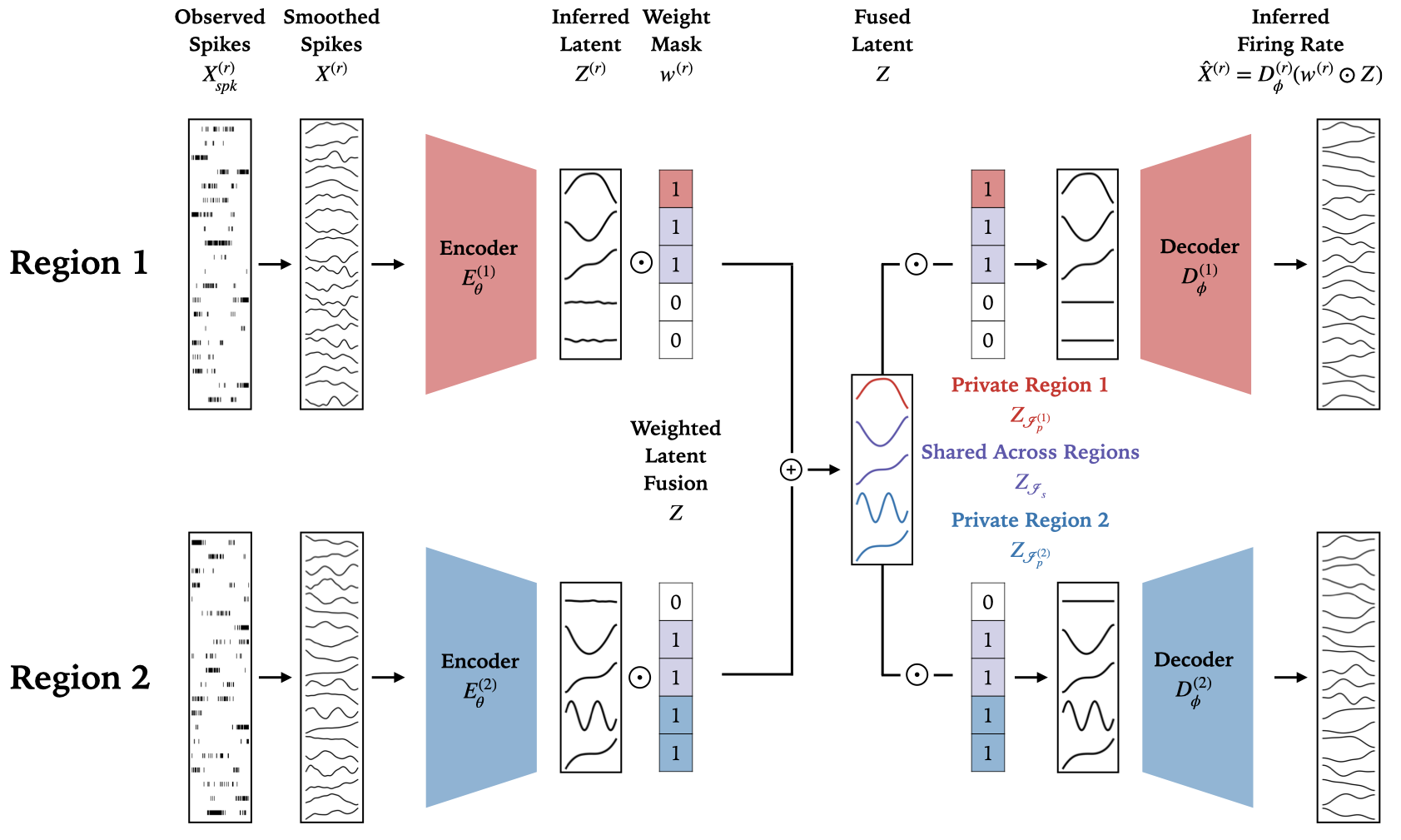}
    \caption{\textbf{\CTAE{} architecture.} \CTAE{} is composed of a coupled autoencoder, where the encoders and decoders are causal transformers designed to reconstruct neural activity for each region $r$. The inputs to the network are estimated spike rates from each region. A weight mask per region $w^{(r)}$ is used to disentangle the shared representation (violet) from the region-specific latents (red and blue) in the encoder outputs $Z^{(r)}$. The latents are recovered via end-to-end training.
    }
    \label{fig:model}
\end{figure}
\noindent \textbf{Weighted latent fusion.}
Broadcasting the masks across time, we aggregate the two latent trajectories
dimension-wise via a masked average for each dimension $d$ in the latent:
\begin{equation}
\mathbf{Z}_t[d]
  \;=\;
  \frac{
         \mathbf{w}_1[d]  \mathbf{Z}^{(1)}_t [d]+ \mathbf{w}_2[d] \mathbf{Z}_t^{(2)}[d]}
       {
         \mathbf{w}_1[d] + \mathbf{w}_2[d]}.
\label{eq:weighted_latent_average}
\end{equation}
Equation \eqref{eq:weighted_latent_average} leaves the private blocks
unchanged (they receive weight 1 from only their own region) 
\begin{equation}
\label{eq:latent_block_extract}
\widehat{\bm{P}}^{(1)}
 \;=\;
  \mathbf{Z}_{\mathcal{I}_1}
  \;=\;
  \mathbf{Z}^{(1)}_{\mathcal{I}_1},\quad
\widehat{\bm{P}}^{(2)}
  \;=\;
  \mathbf{Z}_{\mathcal{I}_2}
  \;=\;
  \mathbf{Z}^{(2)}_{\mathcal{I}_2}.
\end{equation}
while averaging the shared block across regions:
\begin{equation}
\label{eq:shared_block_extract}
\widehat{\bm{S}}   \;=\;
  \mathbf{Z}_{\mathcal{I}_s}
  \;=\;
  \tfrac12\bigl(
    \mathbf{Z}^{(1)}_{\mathcal{I}_s}
    +
    \mathbf{Z}^{(2)}_{\mathcal{I}_s}
  \bigr).
\end{equation}
This design will implicitly force alignment of the shared latents of both regions, via minimizing the reconstruction losses we define in the next section.  

\noindent \textbf{Region-specific decoding.}
Each decoder receives only the latent dimensions relevant to its own
region:
\begin{equation}
\widehat{\mathbf{X}}^{(r)}
 \;=\;
 D_\phi^{(r)}\!\bigl( (\mathbf{w}_r\bm{1}_T^\top) \!\odot\! \mathbf{Z}
 \bigr),
\label{eq:region_recon}
\end{equation}
where $(\mathbf{w}_r\bm{1}_T^\top) \in \mathbb{R}^{D \times T}$ is an outer product such that each weight is duplicated for each dimension along all time points. 
Thus, the element-wise product zeroes out unrelated latents, forcing
decoder $r$ to rely exclusively on the subset of dynamics
meaningful for its region.

This architecture (i) aligns both regions in a common latent space of
dimension $D$, (ii) preserves region-specific structure via fixed
relevance masks, and (iii) enforces cross-region consistency through the
fusion rule in Eq.\,\eqref{eq:weighted_latent_average}.
The loss function balancing reconstruction accuracy and alignment
objectives is detailed in Section~\ref{subsec:loss_two}.
We illustrate our architecture in Fig.~\ref{fig:model}.

\subsection{Training Objective}
\label{subsec:loss_two}
We use four loss functions to recover the shared and private latent representations with \CTAE{}:

\noindent \textbf{Reconstruction loss.}  
The first loss is a reconstruction loss, so that each transformer autoencoder faithfully reproduces its own region’s activity:
\begin{equation}
  \mathcal{L}_{\mathrm{rec}}
  \;=\;
  \sum_{r=1}^{2}
  \bigl\|
    \widehat{\mathbf{X}}^{(r)}
    -
    \mathbf{X}^{(r)}
  \bigr\|_F^{2}.
  \label{eq:loss_recons}
\end{equation}

\noindent \textbf{Shared-only reconstruction.}  
To ensure that \emph{all} 
structure common to both regions is routed into the shared block, we define a loss such that each decoder reconstructs the neural activity from its region using \emph{only} the shared representations.   
Let
\(
  \mathbf{w}^{(s)}
  = \mathbf{w}_1\!\odot\!\mathbf{w}_2
  = [\,\mathbf{1}_{d_s},\mathbf{0}_{d_1},\mathbf{0}_{d_2}]^T
\)
be the intersection mask that selects only the shared dimensions.
The loss is
\begin{equation}
  \mathcal{L}_{\mathrm{shared}}
  \;=\;
  \sum_{r=1}^{2}
  \bigl\|
    D_\phi^{(r)}
    \!\bigl(
      (\mathbf{w}^{(s)}\bm{1}_T^\top) \!\odot\!\mathbf{Z}
    \bigr)
    -
    \mathbf{X}^{(r)}
  \bigr\|_F^{2},
  \label{eq:loss_recons_shared}
\end{equation}
where \(\mathbf{w}^{(s)}\) zeroes out all private coordinates from the inputs to the decoder, thus forcing the model to encode every cross-region regularity inside the shared subspace. Without this constraint, shared information can shift into private subspaces, leading to inaccurate representation.

\noindent \textbf{Alignment loss.}  The shared latents are meant to capture only dynamics common to both regions. To enforce this, we align each encoder’s shared output to their average, ensuring consistency across regions and preventing region-specific variance from leaking
into the shared space:
\begin{equation}
\mathcal L_{\mathrm{align}}
=\sum_{r=1}^{2}\bigl\|
(\mathbf w_r\mathbf 1_T^\top)\odot\mathbf Z
-(\mathbf w_r\mathbf 1_T^\top)\odot\mathbf Z^{(r)}
\bigr\|_F^2.
\label{eq:r_align}
\end{equation}
\noindent \textbf{Orthogonality loss.}  
To encourage each latent coordinate to capture distinct, non-redundant
structure, we penalize correlations between \emph{all} rows of the fused
latent matrix \(\mathbf{Z}\in\mathbb{R}^{D\times T}\).
Let \(\mathbf{G}=\frac{1}{T}\mathbf{Z}\mathbf{Z}^{\top}\) denote the empirical
Gram matrix of latent trajectories, whose off-diagonal entries encode
row-wise correlations. We drive only the \emph{off-diagonal} entries of \(\mathbf{G}\) towards zero,
\begin{equation}
\mathcal{L}_{\mathrm{orth}}
  \;=\;
  \bigl\|
    \mathbf{G}-\operatorname{diag}\!\bigl(\mathbf{G}\bigr)
  \bigr\|_F^{2},
  \label{eq:loss_ortho}
\end{equation}
so that every pair of latent dimensions—shared or private—becomes
approximately orthogonal, promoting global disentanglement.

Finally, the complete objective minimizes the weighted sum of the four losses:
\begin{equation}
\mathcal{L}
 \;=\;
 \mathcal{L}_{\mathrm{rec}}
 +\lambda_{\mathrm{align}}\,\mathcal{L}_{\mathrm{align}}
 +\lambda_{\mathrm{shared}}\,\mathcal{L}_{\mathrm{shared}}
 +\lambda_{\mathrm{orth}}\,\mathcal{L}_{\mathrm{orth}}.
 \label{eq:total_loss}
\end{equation}
The weights
\(\lambda_{\mathrm{shared}},
  \lambda_{\mathrm{align}},\lambda_{\mathrm{orth}}\)
are selected on a held-out validation set. Methodology for training the architecture is summarized in Algorithm~\ref{alg:2rctae}.

\subsection{Beyond two regions}
\label{sec:r3}
For $R\!\ge\!3$, we assign each latent dimension to an arbitrary subset of regions using a binary membership matrix $\mathbf W\!\in\!\{0,1\}^{R\times D}$ with $\mathbf W[r,d]=1$ iff region $r$ uses dimension $d$. The fused latent at time $t$ is a masked average
\[
\mathbf Z_t[d]=\frac{\sum_{r=1}^{R}\mathbf W[r,d]\;\mathbf Z^{(r)}_t[d]}{\sum_{r=1}^{R}\mathbf W[r,d]},
\]
which leaves private dimensions unchanged (claimed by exactly one region) and forces alignment of shared dimensions (claimed by two or more regions). For example, for $R{=}3$, the codes \texttt{100}, \texttt{010}, \texttt{001} denote private latents; \texttt{110}, \texttt{101}, \texttt{011} denote pairwise-shared latents; and \texttt{111} denotes globally shared latents. Complete details and training losses are in Appendix~\ref{sec:ctae_multi}.

\section{Experiments}
\label{sec:experiments}
We analyze two real-world datasets of simultaneous neural recordings from a motor circuit and from a multisensory circuit. 
We demonstrate the efficacy of \CTAE{} in extracting shared and private dynamics in comparison to DLAG~\citep{gokcen2022dlag}. 
Implementation details for each experiment are in the appendix.

\subsection{Motor Circuit: M1-PMd}
\label{sec:motor-main}
We applied \CTAE{} to simultaneous neural recordings from dorsal premotor cortex (PMd) and primary motor cortex (M1) in macaque monkeys performing a standard delayed center-out reaching task with eight outward targets \citep{perich2018neural}, additional details in Appendix~\ref{sec:m1_pmd_data}. The dataset consisted of 208 total trials across 8 reach conditions, with spike-sorted data from 66 and 52 putative neurons in PMd and M1, respectively, recorded via 64-channel Utah arrays. Each trial spanned 3s, with spikes and behavioral variables (e.g., position) binned in 100 ms bins, resulting in 30 time points per trial.

Historically, PMd and M1 have been jointly analyzed due to their complementary roles in movement preparation and execution. During the instructed delay period between stimulus onset and the go cue, both regions exhibit preparatory activity without inducing muscle output \citep{kaufman2014cortical, churchland2024preparatory}, potentially representing task goals\textemdash i.e., target identity \citep{byron2010neural, lara2018conservation}. Following the `go' cue, both PMd and M1 generate execution-related activity that descends to motor neurons and enables decoding of continuous hand kinematics, including position and velocity \citep{churchland2024preparatory, elsayed2016reorganization, gilja2012refit, byron2010neural}.

We applied \CTAE{} to the dataset for unified modeling of joint latent subspaces and identified:
\begin{itemize}
[noitemsep, topsep=0pt]
    \item A shared subspace of dimension $d_s$, capturing correlated dynamics across PMd and M1.
    \item Private subspaces of dimensions $d_{\text{PMd}}$ and $d_{\text{M1}}$, capturing region-specific activity.
\end{itemize}

\begin{figure}[ht]
    \centering
    \includegraphics[width=0.9\linewidth]{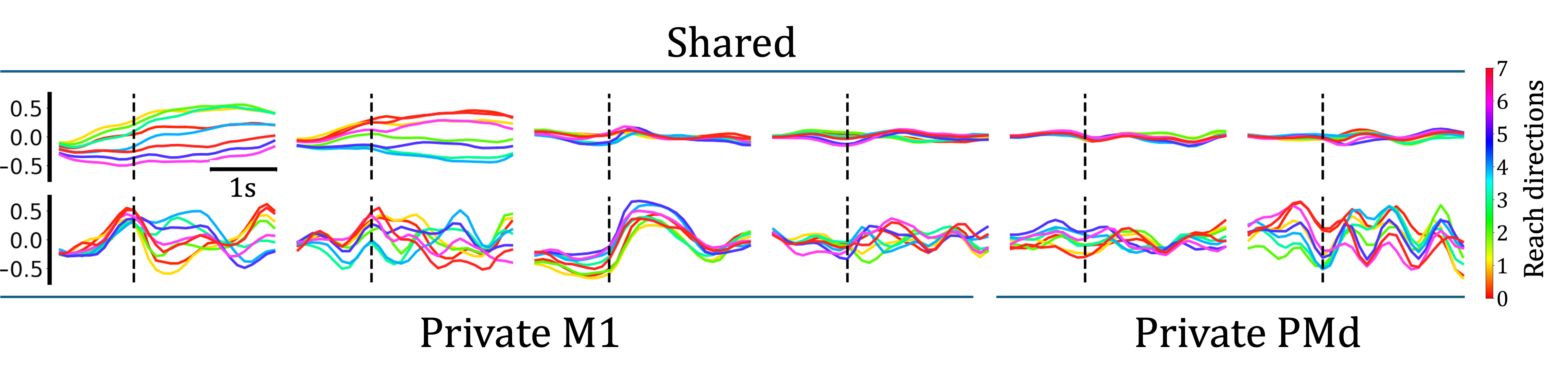}
    \caption{M1-PMd dataset. \CTAE{} shared (top) and region-private (bottom) latents. Dotted vertical line indicates the “go” cue in each trial.    }
    \label{fig:motor2}
\end{figure}

Fig.~\ref{fig:motor2} visualizes a subset of the CTAE inferred neural latent dynamics (the set of all latents is in Fig.~\ref{fig:m1pmd-all-latents}). Each panel shows a distinct latent dimension evolving over time. The shared latent subspace captures both condition-invariant and condition-dependent structure, encoding aspects such as reach direction and temporal progression. Region-specific latents display more complex dynamics, likely reflecting local circuit processes.

\begin{figure}[t!]
    {\centering
    \includegraphics[width=1\linewidth]{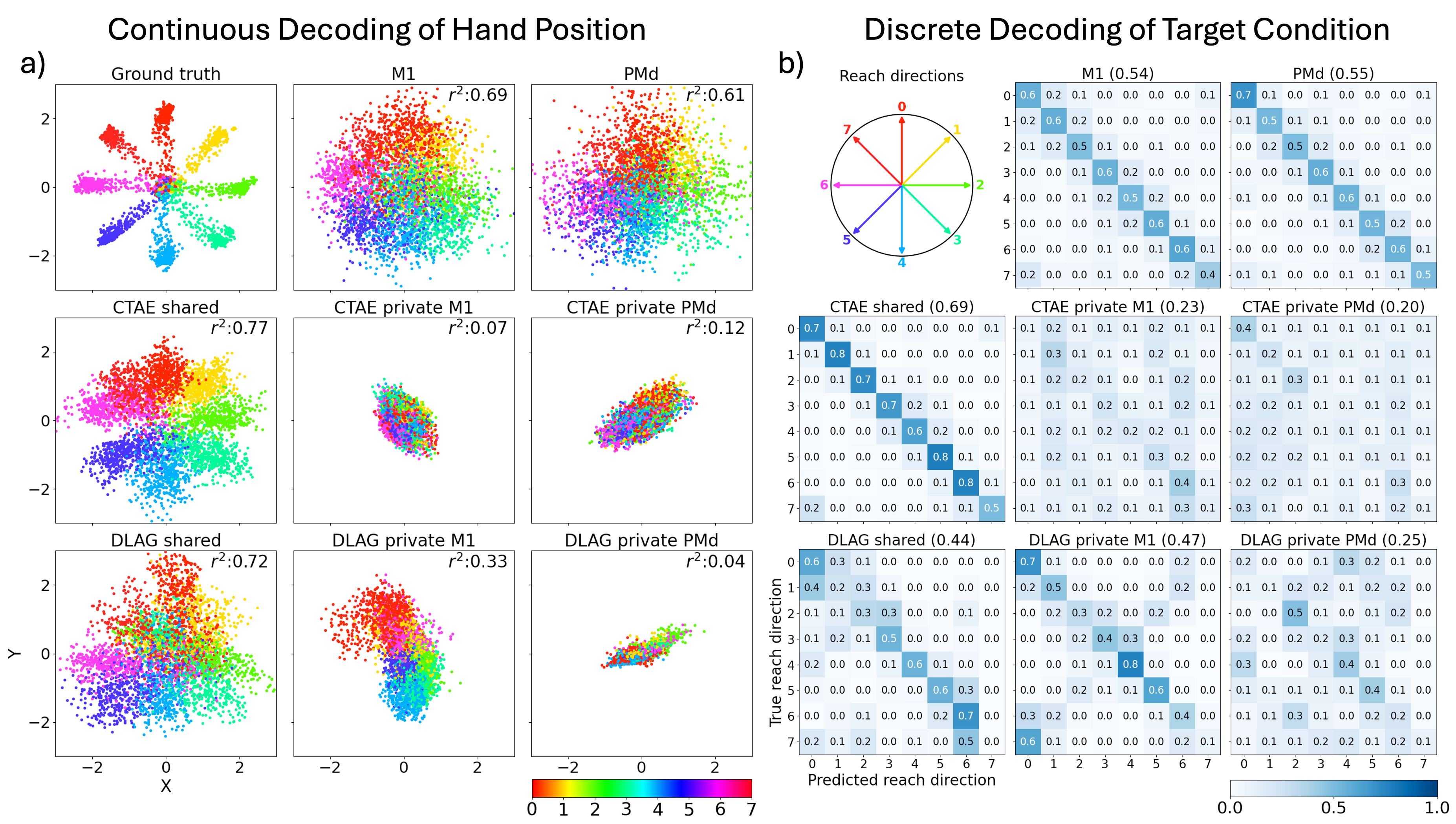}}
    \caption{M1-PMd dataset. 
    a) Ground truth hand position in top-left corner. Hand position decoding from neural activity in M1 and PMd (top), from \CTAE{} latents (middle) and from DLAG (bottom).
    b) Confusion matrices for reach direction classification (order of plots is same as in A). Classification accuracy in parentheses.} 
    \label{fig:motor1}
\end{figure}

To interpret the latent dynamics—shared $Z^{s}$, and private 
$Z^{(\text{PMd})}$, $Z^{(\text{M1})}$, we evaluated the inferred latents using two decoding tasks with simple linear decoders:

\noindent {\bf Continuous decoding hand position}: Predict hand position at each time point after go-cue (last 2 seconds for each trial) using linear regression:
      $
      v_t^{\text{hand}} = W \mathbf{z}_t.
      $
      We plot in Figure~\ref{fig:motor1}A the predictions from neural activity, from \CTAE{} latents and from DLAG latents.  
      
\noindent {\bf Discrete decoding target condition}: Classify the reach condition (one of 8 targets) using multi-class logistic regression:
    $
    \hat{y}_i = \frac{\exp(\mathbf{w}_i^\top \mathbf{z} + b_i)}{\sum_{j=1}^{C} \exp(\mathbf{w}_j^\top \mathbf{z} + b_j)} \quad \text{for } i = 1, \ldots, C.
    $
     We present in Figure~\ref{fig:motor1}B the confusion matrices across conditions for neural activity, \CTAE{} and DLAG latents.

We hypothesized that the shared latent space would encode the majority of behaviorally relevant information—particularly target identity—while PMd might contribute higher-order planning signals, and M1 reflect finer temporal structure related to movement execution.

Our results confirmed that \CTAE{}'s shared latent factors accounted for the majority of variance associated with reaching behavior (Fig. \ref{fig:motor1} a). In other words, the shared PMd–M1 subspace captured the dominant task-relevant signals. In contrast, DLAG tended to assign behaviorally relevant variance to both PMd and M1 private subspaces, potentially due to less effective separation between shared and private dynamics. These findings demonstrate that \CTAE{} can simultaneously decode continuous behavioral trajectories (Fig. \ref{fig:motor1} a). and classify discrete behavioral states (Fig. \ref{fig:motor1} b). more effectively than prior approaches. We present time-wise prediction of the reach in Fig.~\ref{fig:m1pmd_condition_decoding_time}. Importantly, our results support the view that motor planning and execution signals are largely embedded in shared population dynamics across PMd and M1, while private activity may facilitate flexible, context-dependent processing—such as adaptation to perturbations—without interfering with ongoing execution. 

\noindent {\bf Ablation study:} To  evaluate the contribution of the individual loss functions to the CTAE model we perform an ablation study (full details in Appendix~\ref{sec:ablation-overview}) and report the prediction accuracy in Tab.~\ref{tab:model_performance}. 
We demonstrate that removing any of the three loss functions results in decreased performance.

\begin{table}[ht]
    \centering
    \begin{tabular}{lccc}
        \hline
        \textbf{Model} & \textbf{Shared} & \textbf{Private M1} & \textbf{Private PMd} \\ \hline
        CTAE & 0.69 (0.03) & 0.22 (0.02) & 0.21 (0.03) \\
        CTAE without alignment loss & 0.61 (0.02) & 0.16 (0.02) & 0.2 (0.02) \\
        CTAE without orthogonality loss & 0.31 (0.02) & 0.28 (0.02) & 0.29 (0.02) \\
        CTAE without shared only reconstruction loss & 0.34 (0.01) & 0.36 (0.01) & 0.37 (0.01) \\ \hline
        
    \end{tabular}
    \caption{Ablation study of CTAE loss functions. Each entry indicates the model's accuracy in predicting the reach direction. The values in the brackets indicate standard deviation across 5-folds.}
    \label{tab:model_performance}
\end{table}

\begin{figure}[t!]
    \centering
    \includegraphics[width=1.0\linewidth]{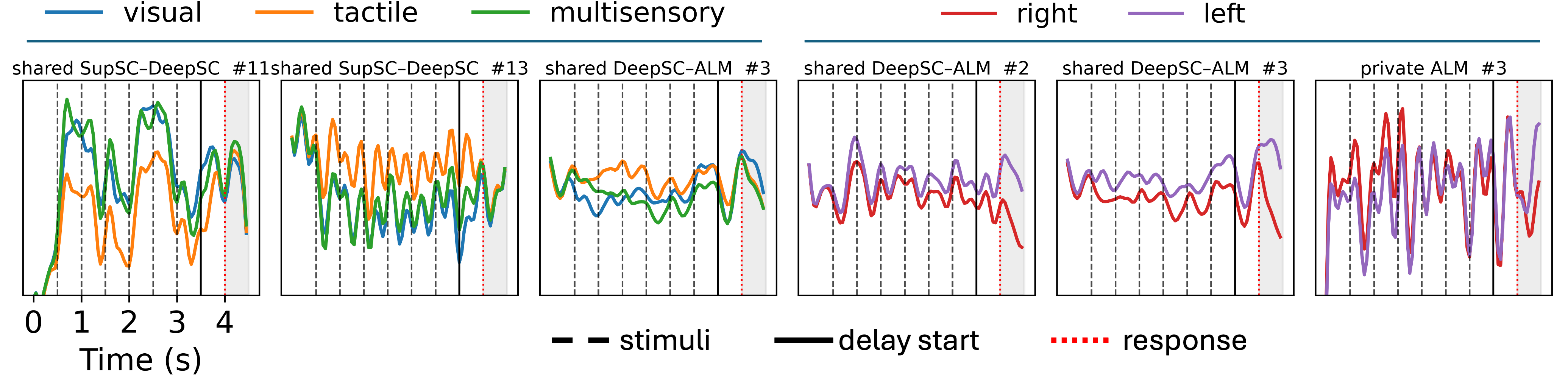}
    \caption{SC-ALM dataset. Representative latents. Each panel shows the condition-averaged time course of one latent (title = subspace and latent index). Left panels: stimuli—visual (blue), tactile (orange), multisensory (green). Right panels: target side—right (red), left (purple). 
    }
    \label{fig:sc_alm_latents}
\end{figure}

\subsection{Multisensory Circuit: SC--ALM}
\label{sec:multisensory-main}
We evaluated CTAE on a new dataset of simultaneous recordings from the superior colliculus (SC) and anterolateral motor cortex (ALM) in mice trained on a multisensory discrimination task (see Appendix~\ref{sec:alm_sc_data} for task details). Using high-density Neuropixels probes, neural activity was recorded from superficial and deep layers of SC together with ALM while animals integrated visual and tactile stimuli to identify the target side (left vs. right). Each trial consisted of six brief, sequential cue epochs (0.5 secs each); at each epoch, a visual, tactile, or multisensory (visual + tactile) stimulus appeared on either the left or right side. The animals were required to report, after a short 0.5 secs delay, the side with the greater number of cues—thus performing a temporal evidence accumulation task. Correct responses were rewarded with water. We fit our multi-region CTAE ($R>2$; Appendix~\ref{sec:ctae_multi}) to this three-region dataset (superficial SC, deep SC, ALM), which enables analysis of shared-across-all, pairwise-shared, and region-private latent interactions (Fig.~\ref{fig:sc_alm_latents}).

SC and ALM are key components of a cortico-subcortical loop implicated in linking sensory inputs to motor planning. SC integrates visual and tactile information and contributes to orienting behavior \citep{Stein1993, cang2013developmental, felsen2008neural}, whereas ALM encodes preparatory and choice-related activity during decision-making tasks \citep{guo2014flow, li2015motor}. We hypothesized that shared subspaces across superficial and deep SC layers and between deep layers of SC and ALM would preferentially capture task-relevant features, such as stimulus type and target side, thereby mediating the transformation from multisensory evidence accumulation to decision making.

\begin{wrapfigure}[27]{r}{0.64\textwidth}
\vspace{-12pt}
    \centering
    \includegraphics[width=1\linewidth]{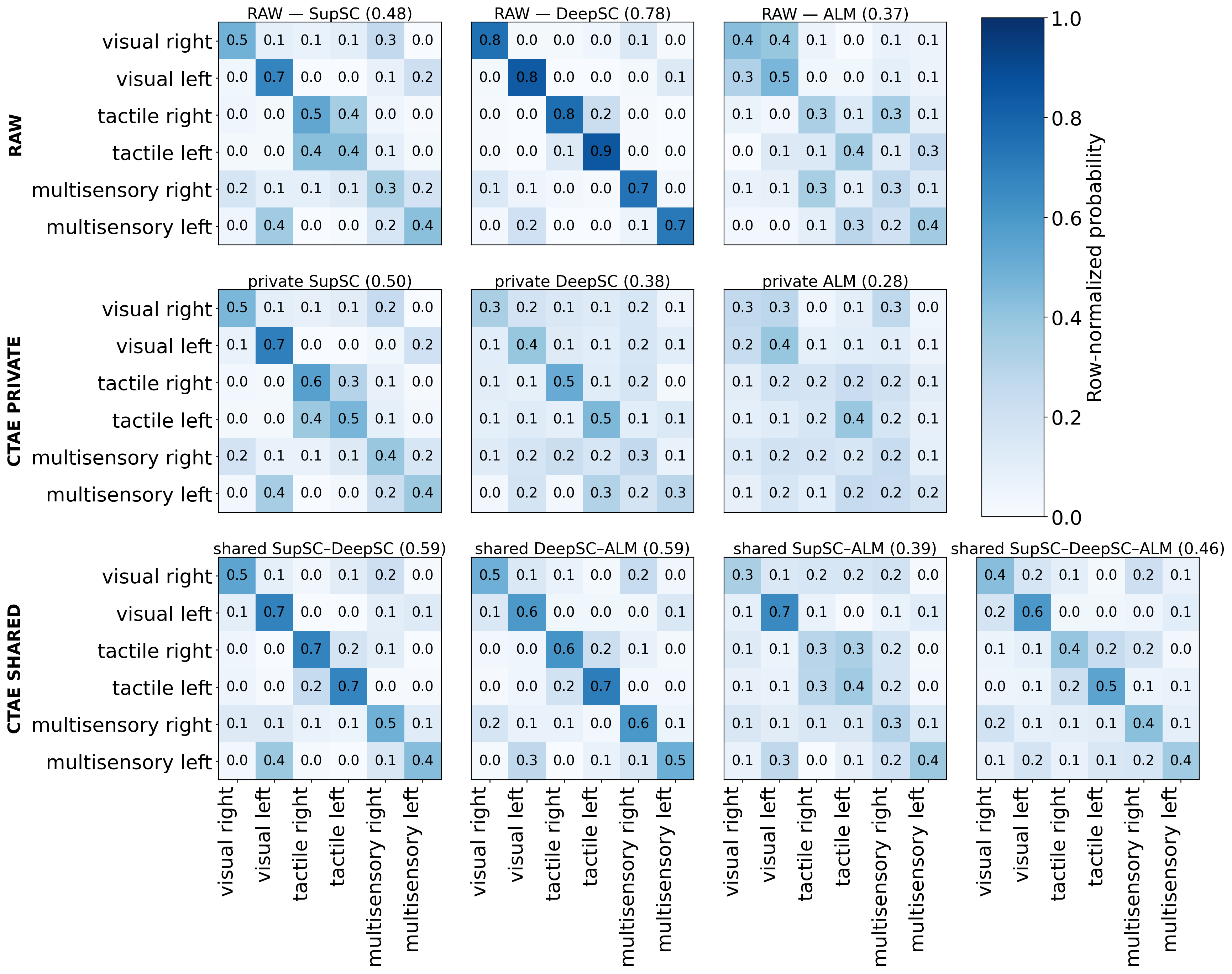}
    \vspace{-20pt}
    \caption{ Row-normalized confusion matrices for decoding stimulus × target side. (top) Raw activity from SupSC, DeepSC, ALM; (middle) CTAE private: region-specific latents; (bottom) CTAE shared latents (pairwise and 3-way). Panel titles show mean 5-fold accuracy.}
    \label{fig:sc_alm_results}
\end{wrapfigure}
Figure~\ref{fig:sc_alm_results} shows decoding performance of stimuli and target side across private and shared subspaces. 
For the raw neural activity, Deep SC shows the strongest overall decoding performance, accurately distinguishing both stimulus type and target side, suggesting its central role in mediating information flow within this circuit.
Superficial SC activity correctly distinguishes target side but is less selective for stimulus modality, whereas ALM activity captures target side more reliably than stimulus type.
Together, these patterns indicate that while all regions carry task information, Deep SC provides the most integrated representation of sensory and choice variables.
With \CTAE{}, shared subspaces between regions reveal clearer and more functionally specific decoding structure.
Shared dynamics between superficial and deep SC, and between deep SC and ALM, yield markedly higher accuracy than region-private latents, reflecting functional coupling along the SC–ALM pathway.
In contrast, the shared subspace between superficial SC and ALM captures little discriminative structure, consistent with the lack of direct anatomical connectivity between these areas.
Overall, these findings suggest that Deep SC serves as a central hub, sharing evidence-related signals with superficial SC and choice-related signals with ALM, supporting the transformation from multisensory integration to motor planning.
Time-resolved decoding further reveals complementary temporal roles across subspaces (Fig.~\ref{fig:sc_alm_decoding}): private latents dominate during the delay and choice periods, maintaining task-relevant information, whereas shared latents ramp in accuracy near stimulus onset and response, reflecting progressive cross-area integration of sensory and decision signals.

\section{Conclusions}
In this paper, we presented a new approach for recovering shared and private latents from multi-region neural recordings. 
Our framework relies on a Transformer-based autoencoder architecture for nonlinear modeling of the relationship between neural activity and latent dynamics, and loss functions designed to recover orthogonal private and shared latents that maximally capture the information in the shared activity between the brain regions.  
By explicitly separating shared and private sources of variability, CTAE facilitates more robust comparisons across contexts and offers a foundation for addressing deeper questions about how distributed neural populations collectively drive flexible behavior and adapt through experience \citep{perich2018neural}.
This work demonstrates effective joint latent inference across cortical areas and establishes a foundation for future studies on generalizable, population-level neural computations across behavioral contexts.

While our paper focuses on neural recordings, our coupled-transformer autoencoder design is not specific to neural data and can be applied to general multiview time series data. 
Our approach is scalable to more than two regions by generalizing the orthogonality loss to pairs of brain regions. Furthermore making the disentanglement weight vector learnable, the dimensions of each of the subspaces can be identified from the training data. These both are directions for future work. 
 
\bibliography{refs}


\newpage

\appendix

\section[CTAE algorithm outline for R=2 brain regions]%
        {CTAE algorithm outline for $R{=}2$ brain regions}
\begin{algorithm}[ht]
\caption{Training Coupled Transformer Autoencoders (\CTAE{}) for two regions} 
\label{alg:2rctae}
\small

\KwIn{Paired recordings $\mathbf{X}^{(1)},\mathbf{X}^{(2)}$, 
 encoder--decoder parameters $\theta,\phi$, 
 latent dimensions $(d_s,d_1,d_2)$,
loss weights $\lambda_{\mathrm{shared}},\lambda_{\mathrm{align}},\lambda_{\mathrm{orth}}$,
optimiser $\mathcal{O}$ and learning rate $\eta$}
\vspace{3pt}
\KwOut{Trained parameters $\theta^\star,\phi^\star$}
\vspace{5pt}

\BlankLine Compute weights:  
$\mathbf{w}_1 \leftarrow \bigl[\mathbf{1}_{d_s},\mathbf{1}_{d_1}, \mathbf{0}_{d_2}\bigr]^\top$, \quad
$\mathbf{w}_2 \leftarrow \bigl[\mathbf{1}_{d_s},\mathbf{0}_{d_1}, \mathbf{1}_{d_2}\bigr]^\top$ \; 
$\mathbf{w}^{(s)} \leftarrow \mathbf{w}_1\odot\mathbf{w}_2$\;

\While{not converged}{
  Encoder forward pass: $\mathbf{Z}^{(1)} \leftarrow E_\theta^{(1)}(\mathbf{X}^{(1)}), \quad
  \mathbf{Z}^{(2)} \leftarrow E_\theta^{(2)}(\mathbf{X}^{(2)})$\;
  Fuse latents: $\mathbf{Z} \leftarrow$ (Eq.~\eqref{eq:weighted_latent_average})\;
  Extract $\widehat{\bm{S}},\widehat{\bm{P}}^{(1)},\widehat{\bm{P}}^{(2)}$ (Eq.~\eqref{eq:latent_block_extract})\;
  Data reconstruction:  
        $\widehat{\mathbf{X}}^{(1)} \leftarrow D_\phi^{(1)}\!\left( (\mathbf{w}_1\bm{1}_T^\top) \!\odot\! \mathbf{Z} \right),
        \quad
         \widehat{\mathbf{X}}^{(2)} \leftarrow D_\phi^{(2)}\!\left( (\mathbf{w}_2\bm{1}_T^\top) \!\odot\! \mathbf{Z} \right)$
        \;
  Compute losses: $\mathcal{L} \leftarrow
    \mathcal{L}_{\mathrm{rec}}
    +\lambda_{\mathrm{shared}}\mathcal{L}_{\mathrm{shared}}
    +\lambda_{\mathrm{align}}\mathcal{L}_{\mathrm{align}}
    +\lambda_{\mathrm{orth}}\mathcal{L}_{\mathrm{orth}} \quad$ (Eq.~\eqref{eq:loss_recons}-\eqref{eq:loss_ortho})   \;

  Parameter update:   $\theta,\phi \leftarrow \mathcal{O}(\theta,\phi,\nabla_{\theta,\phi}\mathcal{L}_{\text{2R}},\eta)$\;
}
\KwRet $\theta^\star,\phi^\star$\;
\end{algorithm}

\section[Extending CTAE to R>2 brain regions]%
        {Extending CTAE to $R{>}2$ brain regions}
\label{sec:ctae_multi}

In the main text, we presented \CTAE{} with a two-region formulation and reported results on datasets involving either two or three regions. Here, we provide the generalization of our framework to $R \geq 3$ regions. 

\subsection{Problem formulation (3 regions)}
\label{sec:problem_3regions}
For clarity, we present the formulation below for three regions. However, this construction naturally extends to more than three regions.

Let $\bm{X}^{(1)} \in \mathbb{R}^{N_{1} \times T}, \bm{X}^{(2)} \in \mathbb{R}^{N_{2} \times T}, \bm{X}^{(3)} \in \mathbb{R}^{N_{3} \times T}$ denote simultaneous neural population recordings from three brain regions, each acquired over $T$ time steps with $N_1, N_2, N_3$ channels, respectively.  
We assume that the observed activity in each region arises from latent dynamics that decompose into:  

\begin{enumerate}
    \item \textbf{Region-private dynamics} $\bm{P}^{(r)}$ that capture computations unique to each region $r \in \{1,2,3\}$, lying in subspaces orthogonal to shared components.  
    \item \textbf{Pairwise shared dynamics} $\bm{S}^{(ij)}$ that span communication subspaces shared between each pair of regions $(i,j) \in \{(1,2), (2,3), (1,3)\}$, reflecting coordinated activity patterns specific to that pair.  
    \item \textbf{Globally shared dynamics} $\bm{S}^{(123)}$ that capture population trajectories common to all three regions, representing fully shared circuit-level computations.  
\end{enumerate}

Formally, the neural activity at each time step $t$ is modeled as nonlinear functions of these latent processes:
\begin{equation}
\begin{aligned}
    \mathbf{X}_t^{(1)} &= f\!\Bigl(\bm{S}^{(12)}_{1:t}, \bm{S}^{(13)}_{1:t}, \bm{S}^{(123)}_{1:t}, \bm{P}^{(1)}_{1:t}\Bigr), \\[6pt]
    \mathbf{X}_t^{(2)} &= g\!\Bigl(\bm{S}^{(12)}_{1:t}, \bm{S}^{(23)}_{1:t}, \bm{S}^{(123)}_{1:t}, \bm{P}^{(2)}_{1:t}\Bigr), \\[6pt]
    \mathbf{X}_t^{(3)} &= h\!\Bigl(\bm{S}^{(13)}_{1:t}, \bm{S}^{(23)}_{1:t}, \bm{S}^{(123)}_{1:t}, \bm{P}^{(3)}_{1:t}\Bigr),
\end{aligned}
\end{equation}
where $\bm{S}^{(ij)}_t \in \mathbb{R}^{d_{ij}}$ denotes the latent state shared between regions $i$ and $j$, $\bm{S}^{(123)}_t \in \mathbb{R}^{d_{123}}$ the fully shared latent state across all three regions, and $\bm{P}^{(r)}_t \in \mathbb{R}^{d_r}$ the region-private latent state of region $r$.  

Our objective is to recover these latent representations $\{\bm{S}^{(ij)}, \bm{S}^{(123)}, \bm{P}^{(r)}\}$ from the observed neural activity $\{\mathbf{X}^{(1)}, \mathbf{X}^{(2)}, \mathbf{X}^{(3)}\}$, such that the model disentangles private, pairwise shared, and fully shared dynamics in a principled way. For more than three regions, this formulation extends by including shared latent variables for all possible combinatorial subsets of regions, with the training algorithm remaining unchanged.
The training algorithm introduced below applies to this general multi-region setting.

\subsection{General multi-region setting}
To move beyond two regions, we extend the formulation with a general mechanism that assigns latent dimensions to all possible combinatorial subsets of regions. This extension builds directly on the same architectural backbone introduced in Section~\ref{sec:two_region}, but augments it with an enhanced weighted latent fusion mechanism. This mechanism enables \CTAE{} to disentangle latents shared by arbitrary subsets of regions while preserving strictly region-private structure. All other components—per-region causal Transformer encoders/decoders and the optimization procedure—remain unchanged. The key modifications are described below.

\noindent \textbf{Encoder outputs.}
For each region \(r\in\{1,\dots,R\}\) we keep a dedicated
encoder–decoder pair $\left(E_\theta^{(r)},D_\theta^{(r)}\right)$ and obtain 
\[
\mathbf Z^{(r)} \;=\; E_\theta^{(r)}(\mathbf X^{(r)})\in\mathbb R^{D\times T}.
\]

\noindent \textbf{Region-specific weight masks.}
A single binary matrix encodes which latent dimensions are associated with each region
\emph{uses}:
\[
\mathbf W =
\bigl[\mathbf w_1\;\mathbf w_2\;\dots\;\mathbf w_R\bigr]^\top
\in\{0,1\}^{R\times D},
\qquad
\mathbf W[r,d]=1
\;\Longleftrightarrow\;
\text{region }r\text{ claims dimension }d.
\]
A dimension is \textit{private} when exactly one row in \(\mathbf W\) is
active and \textit{shared} when two or more rows are active,
thereby accommodating every subset pattern
(for instance, when \(R=3\) the binary code \texttt{110} indicates a
dimension used by regions 1 and 2 but not 3; \texttt{101} corresponds to
regions 1 and 3; and \texttt{111} denotes a dimension shared by all three).
Relative to the two–region mask~\ref{eq:w_def}, \(\mathbf W\) expands the notion of
``shared” from “both regions” to ``arbitary subset of regions”. The membership matrix~\(\mathbf W\) can be (i) fixed a priori from anatomical knowledge, (ii) selected by a hyper-parameter search that allocates an appropriate number of latent dimensions to each subset pattern (analogous to the two-region masks in Eq.~\eqref{eq:w_def}), or (iii) treated as a fully learnable variable and optimised jointly with the rest of the network.

\noindent \textbf{Weighted latent fusion.}
We fuse the region–specific latents with a masked average
\begin{equation}
\mathbf Z_t[d]  \;=\;
\frac{\sum_{r=1}^{R}\mathbf W[r,d]\,\mathbf Z^{(r)}_t[d]}
     {\sum_{r=1}^{R}\mathbf W[r,d]},
\label{eq:r_fusion}
\end{equation}
so that private dimensions stay unchanged, whereas any dimension claimed
by multiple regions is \emph{forced} to align across them.

\noindent \textbf{Region-specific decoding.}
Each decoder receives only the dimensions it owns,
\[
\widehat{\mathbf X}^{(r)} =
D_\phi^{(r)}\bigl((\mathbf w_r\mathbf 1_T^\top)\odot\mathbf Z\bigr),
\]
exactly as in the two–region case.  Hence the extension does not change
model capacity but merely broadens how “shared” information is defined.

\medskip
This design (i) embeds all regions in one latent space of size \(D\),
(ii) preserves region-specific structure through fixed relevance masks,
and (iii) enforces cross-region consistency via
Eq.\,\eqref{eq:r_fusion}.  

\subsection{Training objective}
\label{subsec:loss_multi}

We retain the four losses used for two regions—reconstruction,
shared-only reconstruction, alignment, and orthogonality—while replacing
two binary masks with the matrix \(\mathbf W\).

\noindent \textbf{Reconstruction.}
The reconstruction loss is unchanged, only we now sum over all regions.
\begin{equation}
\mathcal L_{\mathrm{rec}}
=\sum_{r=1}^{R}\bigl\|\widehat{\mathbf X}^{(r)}-\mathbf X^{(r)}\bigr\|_F^2.
\label{eq:r_rec_multi}
\end{equation}

\noindent \textbf{Shared-only reconstruction.}
To guarantee that \emph{all} cross-region regularities live inside
shared dimensions, we mask out every coordinate used by fewer than two
regions.  Let  
\(
\mathbf s = \mathbf 1_{\{\sum_{r}\mathbf W_{r,\cdot}\ge 2\}}\in\{0,1\}^{D}
\)
and \(
\mathbf w^{(s)}_r = \mathbf w_r\odot\mathbf s
\).
We then reconstruct each region from the shared subspace alone:
\begin{equation}
\mathcal L_{\mathrm{shared}}
=\sum_{r=1}^{R}\bigl\|
D^{(r)}_{\phi}\bigl((\mathbf w^{(s)}_r\mathbf 1_T^\top)\odot\mathbf Z\bigr)
-\mathbf X^{(r)}\bigr\|_F^2.
\label{eq:r_shared_multi}
\end{equation}
When \(R=2\) this term reduces to Eq.\,(11) in the main text, but for
\(R>2\) it now supervises \emph{every} shared subset simultaneously.

\noindent \textbf{Alignment.}
Each region’s encoder estimate should coincide with the fused latent
over the dimensions it owns:
\begin{equation}
\mathcal L_{\mathrm{align}}
=\sum_{r=1}^{R}\bigl\|
(\mathbf w_r\mathbf 1_T^\top)\odot\mathbf Z
-(\mathbf w_r\mathbf 1_T^\top)\odot\mathbf Z^{(r)}
\bigr\|_F^2.
\label{eq:r_align_multi}
\end{equation}
Private coordinates contribute zero because they match by definition;
shared coordinates are actively driven together, extending the simple
two-region difference penalty to an $R$-way match.

\noindent \textbf{Orthogonality.}
\begin{equation}
\mathcal L_{\mathrm{orth}}
=\bigl\|
\tfrac{1}{T}\mathbf Z\mathbf Z^\top
-\operatorname{diag}\!\bigl(\tfrac{1}{T}\mathbf Z\mathbf Z^\top\bigr)
\bigr\|_F^2.
\label{eq:r_orth_multi}
\end{equation}
Identical to the two-region version, this term keeps all latent
dimensions—private or shared—mutually decorrelated.

\noindent \textbf{Total loss.}
\begin{equation}
\mathcal L
=\mathcal L_{\mathrm{rec}}
+\lambda_{\mathrm{shared}}\mathcal L_{\mathrm{shared}}
+\lambda_{\mathrm{align}}\mathcal L_{\mathrm{align}}
+\lambda_{\mathrm{orth}}\mathcal L_{\mathrm{orth}},
\end{equation}
with \(\lambda_{\mathrm{shared}},\lambda_{\mathrm{align}},
\lambda_{\mathrm{orth}}\) selected on a validation set.  The objective
reduces exactly to the two-region formulation when \(R=2\); for
\(R>2\) it differs only in how masks are constructed and applied,
maintaining conceptual continuity while capturing richer patterns of
shared neural dynamics.

\section{Training and Hyperparameter Details}
\label{app:training}

To ensure consistency in model capacity across regions, we adopt identical causal Transformer-based encoder–decoder architectures for each region, using an equal number of encoder and decoder layers to maintain architectural symmetry. We then performed a grid search over:

\noindent \textbf{Number of layers.} 
We varied the total number of Transformer layers $L\in\{1,2,3\}$ in both encoder and decoder to trade off model capacity against overfitting risk and computational cost.

\noindent \textbf{Latent dimensions.} Private and shared dimensions, $d\in\{5,10,15\}$.
Note - Here, the specified latent dimensionality $d$ should be treated as an upper bound on bottleneck capacity. After training with the designed regularizers, redundant factors collapse to near-zero explained variance, yielding an effective dimensionality $d_{\text{eff}}\le d$ that approximates the data’s intrinsic dimensionality.

\noindent \textbf{Loss weights.}
    \begin{itemize}
      \item Shared‐space: $\lambda_{\mathrm{shared}}\in\{0,1,2\}$
      \item Alignment: $\lambda_{\mathrm{align}}\in\{0, 0.05, 0.1, 0.5\}$
      \item Orthogonality: $\lambda_{\mathrm{orth}}\in\{0,0.001,0.01,0.05\}$
    \end{itemize}
\noindent \textbf{Learning rate.} $\eta\in\{10^{-3}, 10^{-4}\}$

\noindent \textbf{Warm-up schedule for orthogonality loss.}  
  To prevent an overly strict orthogonality constraint from hindering early representation learning, we set 
  $\lambda_{\mathrm{orth}}=0$ for the first $e$ epochs—allowing the model to focus on reconstruction and shared/private separation—then linearly ramp it up to its target $\lambda_{\mathrm{orth}}$ value over the next $e$ epochs.  This gradual increase stabilizes training by delaying the full orthogonality penalty until the autoencoders have already learned meaningful features.  We explored $e\in\{100,500,1000\}$.  Formally,
    \[
      \lambda_{\mathrm{orth}}^{(t)} =
      \begin{cases}
        0, & t \le e,\\
        \displaystyle\frac{t - e}{e}\,\lambda_{\mathrm{orth}}, & e < t \le 2e,\\
        \lambda_{\mathrm{orth}}, & t > 2e\,. 
      \end{cases}
    \]
  \noindent \textbf{Positional encoding.}  
  We use fixed, deterministic sinusoidal signals into each input embedding to convey temporal order, following~\citep{vaswani2017attention}.  Specifically, for time step $t$ and embedding dimension index $i$ (with model dimension $d_{\mathrm{model}}$):
  \[
    \mathrm{PE}(t,2i)   = \sin\!\Bigl(\tfrac{t}{10000^{2i/d_{\mathrm{model}}}}\Bigr), 
    \quad
    \mathrm{PE}(t,2i+1) = \cos\!\Bigl(\tfrac{t}{10000^{2i/d_{\mathrm{model}}}}\Bigr).
  \]
  These encodings provide both absolute and relative position information without adding any learnable parameters, enabling the self-attention layers to distinguish different time steps.

All models were trained for 10\,000 epochs, and the run minimizing the total loss in \eqref{eq:total_loss} on a held‐out validation set was selected.  Table~\ref{tab:hyperparam_search} summarizes all the hyperparameters and their corresponding search ranges and Table \ref{tab:best_params} reports the optimal settings for the two datasets studied.

\begin{table}[ht]
  \centering
  \caption{Hyperparameter search ranges.}
  \label{tab:hyperparam_search}
  \begin{tabular}{lcc}
    \toprule
    \textbf{Hyperparameter} & \textbf{Values} \\
    \midrule
    \# Transformer layers $L$ & $\{1,2,3\}$ \\
    latent dims.\ $d_1 = d_2 = d_s$ & $\{5,10,15\}$ \\
    $\lambda_{\mathrm{shared}}$     & $\{0,1,2\}$ \\
    $\lambda_{\mathrm{align}}$      & $\{0,0.05,0.1,0.5\}$ \\
    $\lambda_{\mathrm{orth}}$       & $\{0,0.001,0.01,0.05\}$ \\
    Learning rate $\eta$            & $\{10^{-3}, 10^{-4}\}$ \\
    Warm‐up epochs for orth.\ loss  & \{100,500,1000\} \\
    \bottomrule
  \end{tabular}
\end{table}

\begin{table}[ht]
  \centering
  \caption{Selected hyperparameters for M1–PMd and SC-ALM.}
  \label{tab:best_params}
  \begin{tabular}{lcc}
    \toprule
    \textbf{Parameter} & \textbf{M1–PMd} & \textbf{SC–ALM} \\
    \midrule
    latent dims  & $10$ & $15$ \\
    \# Layers $L$           & $3$    & $1$    \\
    $\lambda_{\mathrm{shared}}$ & $1$    & $1$    \\
    $\lambda_{\mathrm{align}}$  & $0.5$  & $0.1$    \\
    $\lambda_{\mathrm{orth}}$   & $0.01$ & $0.01$\\
    Learning rate $\eta$     & $10^{-4}$ & $10^{-4}$ \\
    Warm‐up (orth.\ loss)    & 100 epochs & 1000 epochs \\
    \bottomrule
  \end{tabular}
\end{table}

\noindent \textbf{Compute Resources.}
All models were trained using CUDA-accelerated implementations in PyTorch. Training was performed on NVIDIA Quadro RTX 8000 GPUs, supplemented by additional compute resources from a high-performance computing (HPC) cluster.

\section{Dataset Details}
\label{sec:appendix_task}

The motor circuit dataset used in this study is publicly available and was originally released with \citep{perich2018neural}. Both datasets were preprocessed using standard neural signal processing pipelines. Specifically, spike trains were first binned at 100 ms for both M1–PMd dataset and SC-ALM dataset. The binned spike counts were then smoothed using a Gaussian filter (kernel size of 3 and 2 for M1-PMd and SC-ALM, respectively) to estimate instantaneous firing rates, which were used as input to the CTAE model.

\subsection{Motor Circuit Dataset: Center-Out Arm Reaching Task}
\label{sec:m1_pmd_data}
\begin{figure}[ht]
    \centering
    \includegraphics[width=0.85\linewidth]{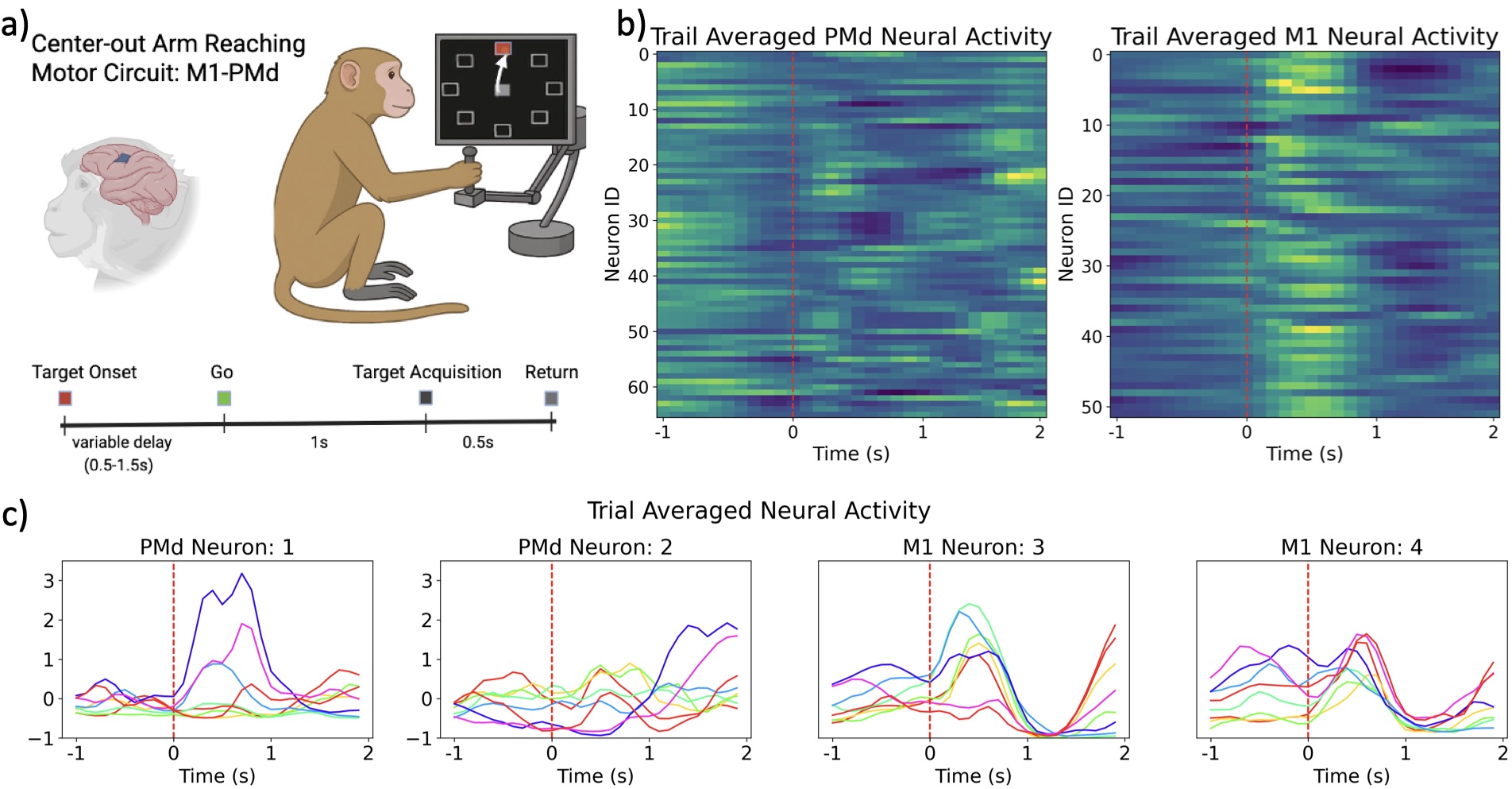}
    \caption{Motor study. a) Schematic of the behavioral tasks performed by non-human primates in the datasets used in Section \ref{sec:experiments}. Center-Out Arm Reach Task: The monkey controls a manipulandum to perform instructed delay reaches toward one of eight peripheral targets, with neural activity recorded from M1 and PMd using Utah arrays. b) Population firing rates in M1 and PMd. Heatmaps shows the trial-averaged z-scored firing rates for individual neurons (rows) across time. c) Condition-averaged firing rates of individual neurons in M1 and PMd. Each panel shows trial-averaged firing rates for single neurons, grouped and colored by reach direction. Colormap follows that in Fig. \ref{fig:motor1}.}
    \label{fig:task_illustration}
\end{figure}

We use the dataset from 
\citep{perich2018neural}, which includes neuronal recordings from a Utah Array in M1 and PMd; see Fig. \ref{fig:task_illustration}, Left). In this task, a monkey performs center-out arm reaches using a planar manipulandum. Neural activity was recorded from the primary motor cortex (M1) and dorsal premotor cortex (PMd) via Utah arrays, yielding spike-sorted activity from 52 putative units in M1 and 66 in PMd (Fig. \ref{fig:task_illustration} Right). Each trial begins with a target onset, followed by a variable delay period (0.5–1.3 s) for movement preparation. After a “go” cue, the monkey initiates a reach to one of eight evenly spaced targets in a 2D workspace, holds briefly, and returns to center. The dataset includes 208 trials across 8 target conditions, with approximately 26 repetitions per condition.

\subsection{SC-ALM Dataset: Multisensory Task} 
\label{sec:alm_sc_data}
The anterolateral motor cortex (ALM) and the superior colliculus (SC) are important parts of a cortico-subcortical loop that transforms multisensory inputs into behavioral decisions. To study the role of these areas in multisensory integration and decision-making, we trained mice in a multisensory discrimination task, where animals had to integrate visual and tactile information over time to identify the target stimulus side. We then performed simultaneous neural recordings in ALM and SC, using high-density Neuropixels probes, in task-performing animals.

\begin{figure}[ht]
    \centering
    \includegraphics[width=0.9\linewidth]{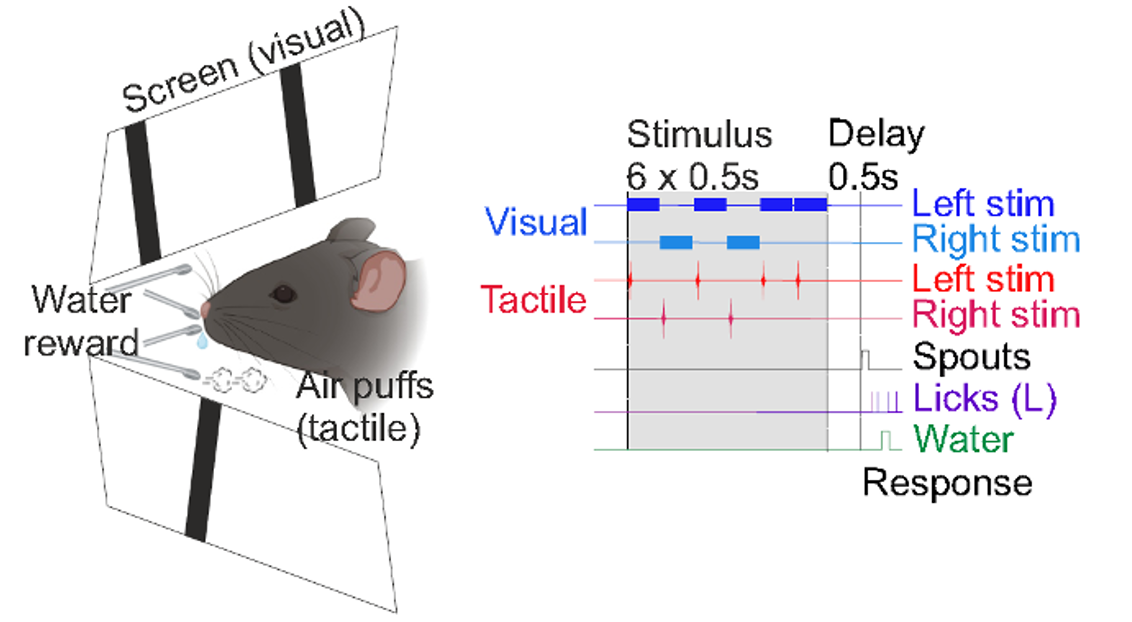}
    \caption{Multisensory study. Left: Image of a mouse in the setup during the response phase. Middle: Schematic of the 2-alternative forced choice task performed by the mice, trained to lick on the side where more stimuli was presented. The mice received a water reward through the target spout.  Right: Schematic of the task timeline in a multisensory trial with a left-side target. The stimulus phase is in grey, the delay and response phase are in white, separated by a vertical line and the movement of the spouts begins at the start of the response period.}
    \label{fig:alm_task_illustration}
\end{figure}

Each trial had a total duration of 7 seconds, comprising a 1-second baseline, then 3 seconds of probabilistic stimuli, followed by a 0.5-second delay without any stimulus and finally, the spouts moved in front of the mouse, allowing up to 2.5 seconds for a licking response. The mice were rewarded when responding on the side where more sensory stimuli were presented during the stimulus period. The stimulus period was composed of six periods of 500 milliseconds each, during which a stimulus could be presented to the mouse. One to six stimuli were presented on a given side, with each stimuli likelihood of presentation determining the difficulty of the task. Depending on the trial type, a stimulus could either be a \emph{visual} grating moving along a screen (visual trial), an \emph{air puff} of 100ms to the vibrissae on one side of the mouse (tactile trial), or both, in a simultaneous congruent \emph{multisensory} manner (multisensory trial). Mice were trained to respond by licking the side on which more stimuli were shown.
An example trial is illustrated in Fig.~\ref{fig:alm_task_illustration}(center).
An example of a multisensory trial timeline is presented in Fig.~\ref{fig:alm_task_illustration}(right).

We analyzed a single session from an expert animal that achieves $\sim 70\%$ correct choice accuracy (564 trials in total), with approximately 80–100 trials per stimulus type and per target direction

\section{Ablation Study: Individual Loss Contributions}
\label{sec:ablation-overview}
We evaluate the contribution of each loss term by training four CTAE variants:
\begin{enumerate}
  \item \textbf{Full CTAE}: all losses active (shared‐only reconstruction, alignment, orthogonality, standard reconstruction).
  \item \textbf{No Shared‐only Reconsruction}: drop the loss that reconstructs inputs using only the shared subspace.
  \item \textbf{No Alignment}: remove the term that aligns shared latents across regions.
  \item \textbf{No Orthogonality}: remove the orthogonality constraint between shared and private latents.
\end{enumerate}
For each variant, we (i) visualize ablated interactions where meaningful, and (ii) quantify reach‐direction decoding accuracy from each latent subspace.

\subsection{Shared‐Only Reconstruction Loss Ablation}
\label{sec:ablate-recon-shared}
The shared‐only reconstruction loss ensures that the shared subspace alone can reconstruct the original neural input; it encourages capturing all common dynamics in that subspace.  Without it, information shifts into private latents, reducing interpretability of the shared space. As shown in Table~\ref{tab:model_performance}, removing this loss component leads to a dramatic drop in shared‐latent decoding accuracy, accompanied by a corresponding increase in private‐latent accuracies.

\subsection{Alignment Loss Ablation}
\label{sec:ablate-align}
\begin{figure}[ht]
    \centering
    \includegraphics[width=0.99\linewidth]{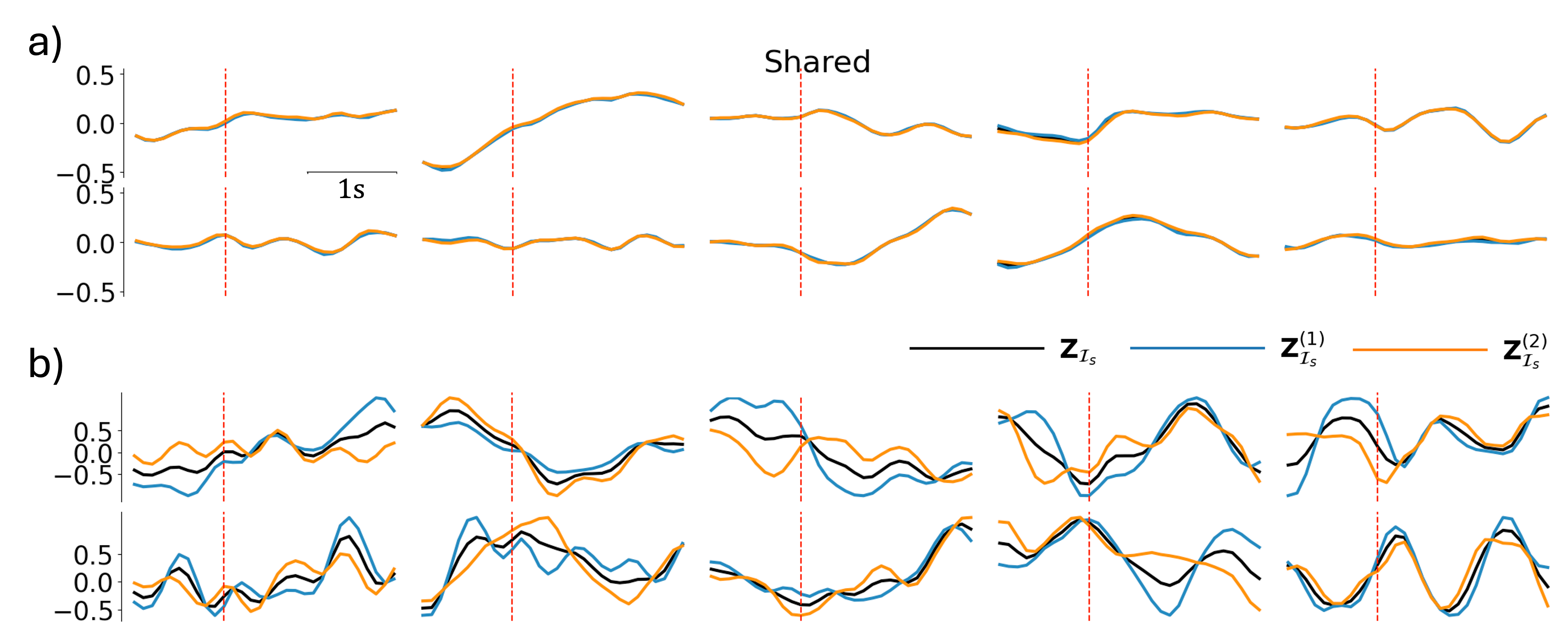}
    \caption{Alignment of shared latents in M1-PMd. Each subplot visualizes a single latent. Blue and orange traces indicate shared latents obtained from M1 and PMd respectively. Black trace represents the mean of blue and orange traces. (a) Alignment loss is enforced in the model. The shared latents have significant overlap. (b) Alignment loss is relaxed in this model. The latents deviate significantly from each other.}
    \label{fig:shared-alignment}
\end{figure}

The alignment loss penalizes discrepancies between shared latents from each region, enforcing that they capture the same underlying dynamics.  
To highlight the significance of this term, we performed loss ablation experiments and analyzed the latents when no alignment loss was enforced.  In Fig. \ref{fig:shared-alignment}a), the alignment loss is enforced, forcing the latents to have high similarity and significant overlap.  When the alignment loss is relaxed in Fig. \ref{fig:shared-alignment}b), the two shared latents diverge and sometimes cancel each other, indicating a breakdown in cross‐region correspondence.  Correspondingly, Table~\ref{tab:model_performance} shows that removing the alignment loss degrades shared‐latent decoding accuracy, while private‐latent accuracies remain largely unchanged.

\subsection{Orthogonality Loss Ablation}
\label{sec:ablate-ortho}
\begin{figure}[ht]
    \centering
    \includegraphics[width=0.9\linewidth]{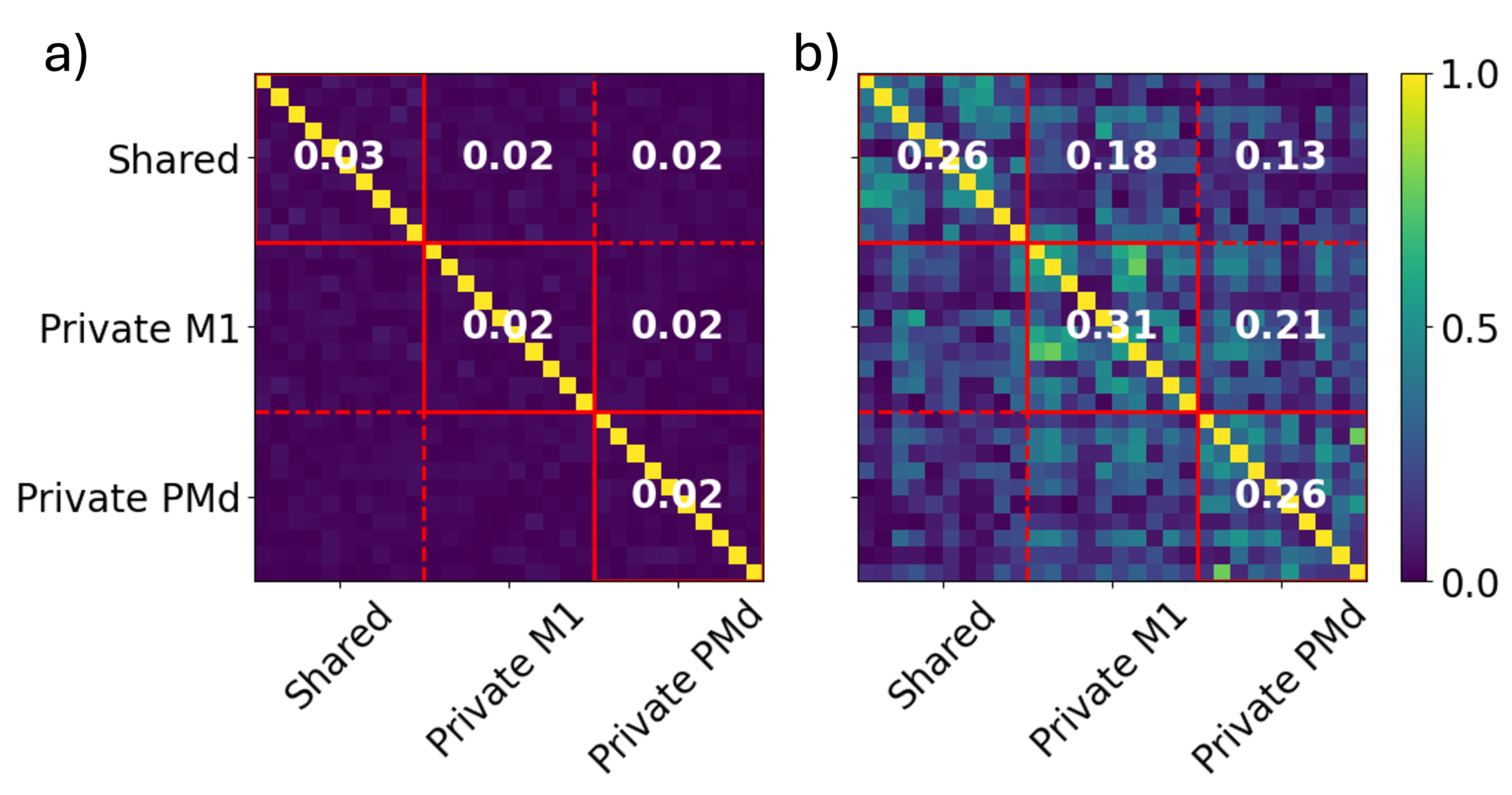}
    \caption{Orthogonality between CTAE latents in M1-PMd. The red lines are used to differentiate between the shared and region-private M1 and region-private PMd latents, in that order. The values in the blocks indicate the mean of dot product value between the latents. (a) Latents from model with orthogonality loss enforced (b) Latents from model with orthogonality loss relaxed}
    \label{fig:ortho-latents}
\end{figure}

The orthogonality loss enforces that shared and private latents carry non‐redundant information by minimizing their dot products.  
 To evaluate the importance of this loss, ablation experiments with relaxed orthogonality constraints were used during model training. The degree of disentanglement is quantified using the dot products between latents. In Fig. \ref{fig:ortho-latents} a) the orthogonality loss is enforced in the model and this constrained the latents to be maximally disentangled from each other. In Fig. \ref{fig:ortho-latents} b) the orthoginality loss is relaxed in the model and there is a significant overlap between the latents, indicating a breakdown in disentanglement. Correspondingly, Table~\ref{tab:model_performance} shows that removing the orthogonality loss degrades shared‐latent decoding accuracy, while private‐latent accuracies remain largely unchanged.

\section{Analysis of latents}
\subsection{Overview of Latent Dynamics}
\label{sec:appendix_latents}
\begin{figure}[ht]
    \centering
    \includegraphics[width=0.99\linewidth]{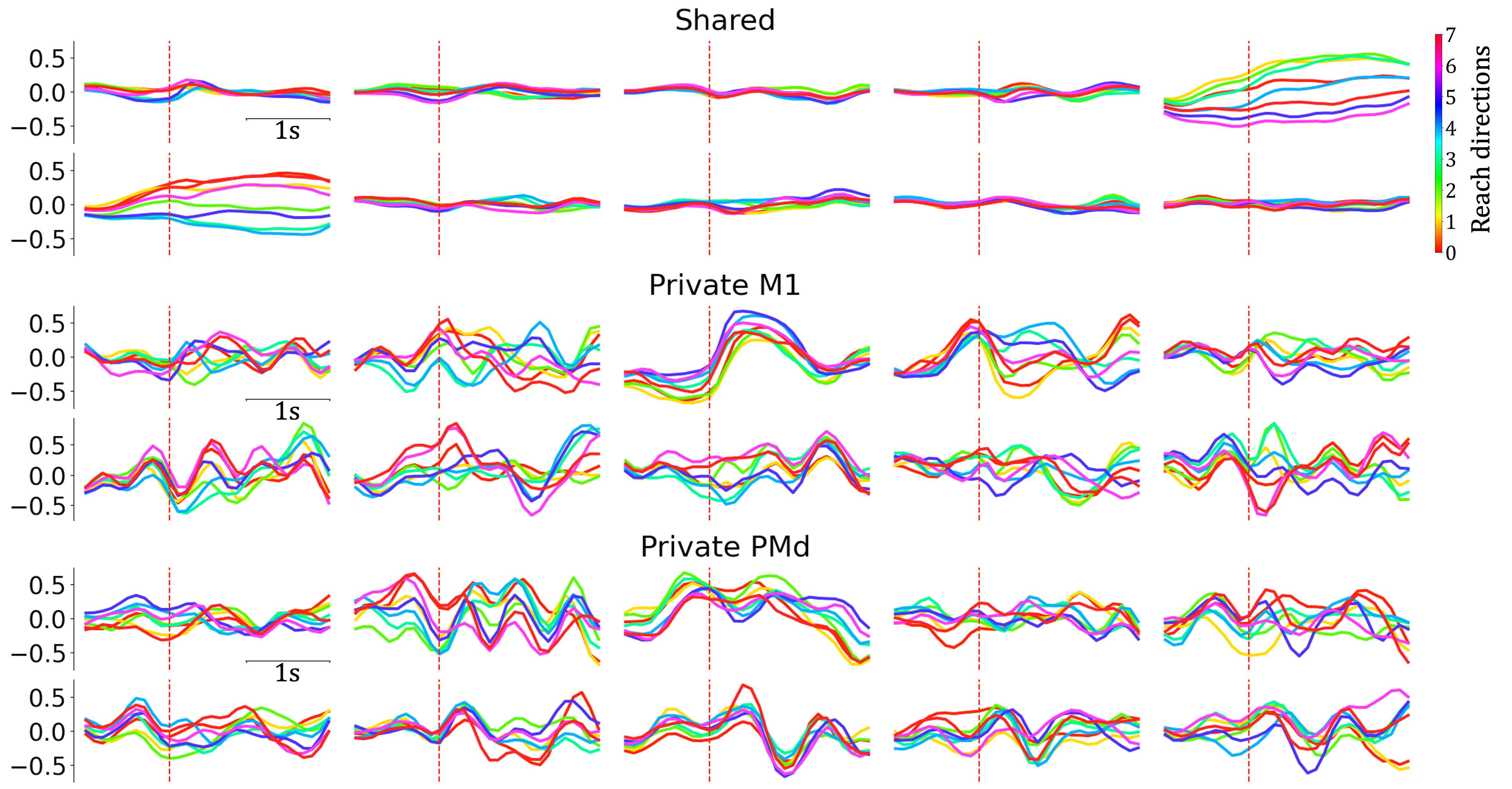}
    \caption{CTAE latent dynamics in M1-PMd. Each subplot visualizes a single neural latent, with colors indicating trial-averaged latents for each reach direction. Red dashed line indicates go-cue. (Top) Shared latents capture neural dynamics present in both M1 and PMd. (Middle) M1-private latents capture dynamics specific to M1 (Bottom) PMd-private latents capture dynamics specific to PMd.}
    \label{fig:m1pmd-all-latents}
\end{figure}

The CTAE model extracts both shared and region-private latents from neural activity, and does so at a single-trial resolution. Shared and region-private latents for the motor dataset are obtained from the best performing CTAE model and visualized in Fig. \ref{fig:m1pmd-all-latents}. The region-private latents capture neural activity patterns that are specific to the region and thereby, reflect local dynamics and computations. The shared latents capture activity patterns that are common to both regions. The representation of task variables such as reach direction in motor task was analyzed by averaging latent trajectories across trials grouped by task variables. In both regions in Fig. \ref{fig:m1pmd-all-latents}, distinct latents exhibit systematic variation across the task variables, indicating that the latents are inherently capturing behaviorally relevant information.

\begin{figure}[ht]
    \centering
    \includegraphics[width=0.9\linewidth]{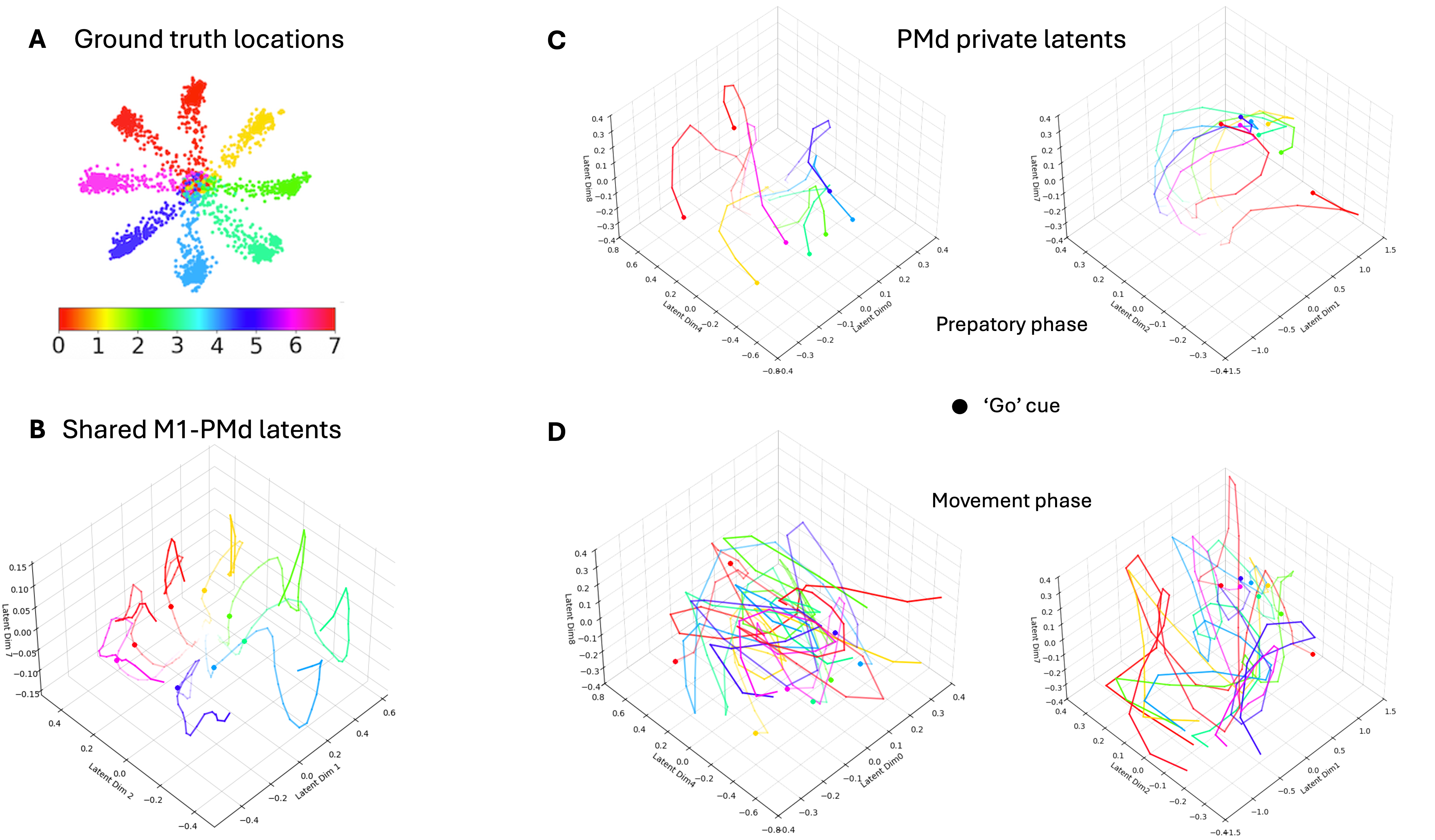}
    \caption{State-space views of CTAE latents across task epochs.
Each subplot shows trajectories in a 3 dimensional latent subspace. Colors denote trial-averaged trajectories for each reach direction in (A). (B) Shared PMd–M1 subspace: trajectories evolve over the entire trial, circle denotes `go' cue. (C) PMd-private subspace during the preparatory phase. Left and right plots are two different 3D subspaces within the private latents. (D) PMd-private subspace during the movement phase. Together, these panels highlight distinct PMd–M1 interaction modes that differ between preparation and execution. Left and right same as in (C).}
    \label{fig:mdpmd_statespace}
\end{figure}

Fig.~\ref{fig:mdpmd_statespace} visualizes a subset of the CTAE inferred neural latent dynamics (from the set of all latents in Fig.~\ref{fig:m1pmd-all-latents}). Each panel shows state-space trajectories for different latent groups across task epochs. Immediately after target onset, trajectories in the shared subspace $Z^{(s)}$ diverge from a common baseline toward target-dependent fixed points, yielding an explicit goal code—consistent with prior reports of a PMd–M1 preparatory subspace. During movement, variance in $Z^{(s)}$ increases along axes orthogonal to the preparatory subspace, indicating a shift in PMd–M1 coupling as the task transitions to execution (Fig.~\ref{fig:mdpmd_statespace}B). 
We also observe that PMd-private latents maintain a stable target representation during the preparatory phase (Fig.~\ref{fig:mdpmd_statespace}C), consistent with staging signals that are subsequently reflected in the shared space; after the go cue, this private structure weakens, no longer reflecting clear separation of different targets, consistent with transient reorganization or monitoring as movement unfolds (Fig.~\ref{fig:mdpmd_statespace}D).

\begin{figure}[t]
  \centering
  \includegraphics[angle=90, origin=c, width=0.28\textheight]{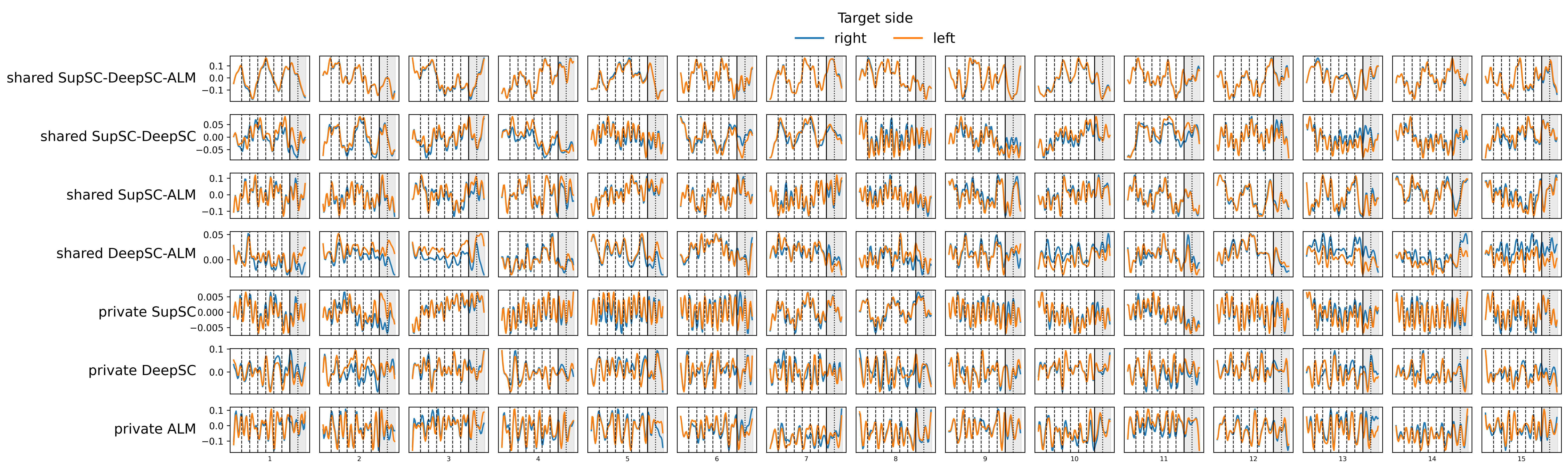}
  \caption{SC-ALM: CTAE latents averaged across target side.}
  \label{fig:sc_alm_target}
\end{figure}

\begin{figure}[t]
  \centering
  \includegraphics[angle=90, origin=c, width=0.28\textheight]{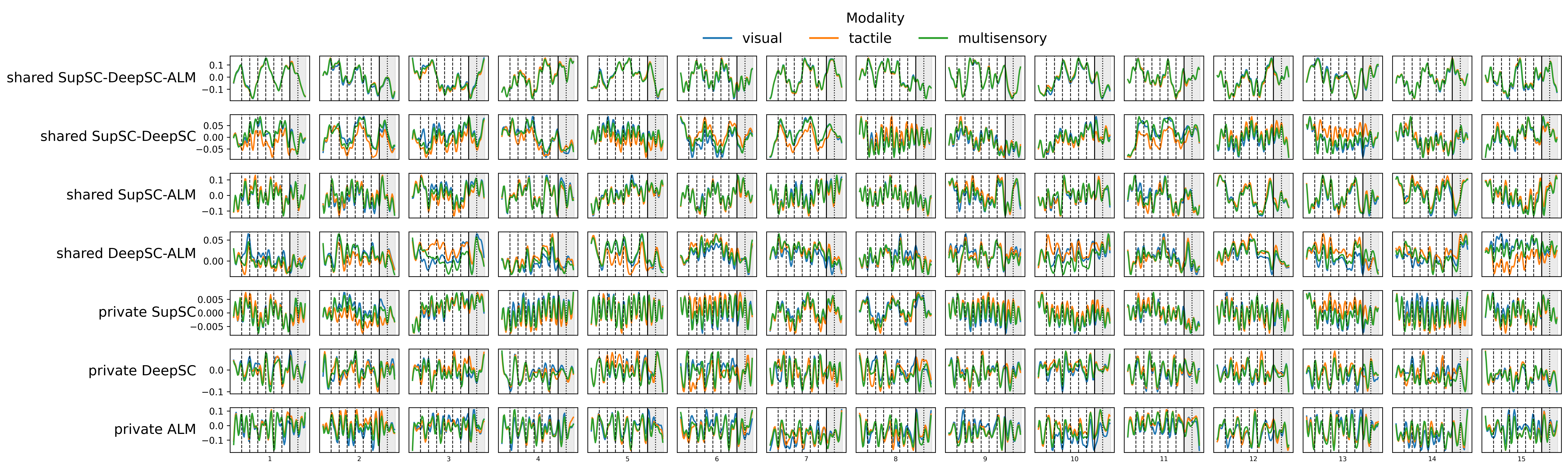}
  \caption{SC-ALM: CTAE latents averaged across stimulus type.}
  \label{fig:sc_alm_stimulus}
\end{figure}

In Fig.~\ref{fig:sc_alm_target} and Fig.~\ref{fig:sc_alm_stimulus}, we present all the latents for the SC-ALM dataset, where we trial-average the latents within target side and stimulus type, respectively.  
In general, most latents exhibit oscillatory patterns that, to varying degrees, are tuned to the stimulus presentation times; superficial SC latents tend to exhibit higher-frequency oscillations. 
Shared latents for Deep SC and ALM, and region-specific ALM latents, exhibit separation between the target sides at the delay and response.
Superficial SC latents and shared superficial and deep SC latents exhibit separate activity pattern for tactile trials compared to visual and multisensory trials. In contrast, Deep SC and ALM shared latents exhibit patterns that separate all three stimulus types. 

We next evaluate temporal decoding of the target variable (e.g., reach direction, target side) from the latents, to better determine the dynamics represented by the latents in shared and private subspaces.

\clearpage

\subsection{Temporal Decoding Analysis from CTAE Latent Subspaces }
\label{sec:temporal_decoding}

\paragraph{Time-resolved decoding.}
To localize when task information appears in each CTAE latent subspaces, we considered per-trial, per-time bin latents from CTAE's private and shared (pairwise and all-shared) subspaces. 
At every time bin, the input features are the latent dimensions in that subspace (i.e., private, all shared, or pairwise shared) within a window of 5 time bins.  
For each subspace  we ran 5-fold stratified cross-validation with a single multinomial logistic-regression classifier (with feature standardization) trained once per fold on the full time-flattened feature matrix. Time-resolved accuracy was then obtained without re-training: at test time we fixed the classifier and masked all features outside a temporal context centered at each time point (a symmetric sliding window of five bins, i.e., $\pm 2$ bins). We report mean $\pm$ s.d. across folds at every time bin and overlay event markers for cue onsets, delay start, and delay end.

\paragraph{Findings: SC-ALM dataset.}
Across subspaces, the accuracy of the shared latents progressively improves as the response approaches, indicating that cross-area information integrates toward the decision and movement execution. 
Specifically, the shared subspace across all 3 regions demonstrates a positive trend towards the response period with the highest accuracy at the response period and a slight increase in accuracy after each stimulus time.   
Superficial SC private subspace shows the highest prediction accuracy prior to the delay period, consistent with sustained maintenance of task-relevant signals. In contrast, ALM and the Deep SC–ALM shared subspace peak after the response, suggesting strong encoding of the target side and decision (the two are highly correlated in expert mice). The Superficial SC–Deep SC shared subspace also exhibits marked prediction increases immediately following most stimulus onsets, pointing to early multisensory interactions between superficial and deep SC layers, and shared representation of stimulus direction. Together, these patterns reveal complementary temporal roles: private subspaces emphasize delay-period maintenance and response choice, while shared subspaces carry stimulus- and response-proximal information that ramps as action nears.

\begin{figure}[ht]
    \centering
    \includegraphics[width=0.6\linewidth]{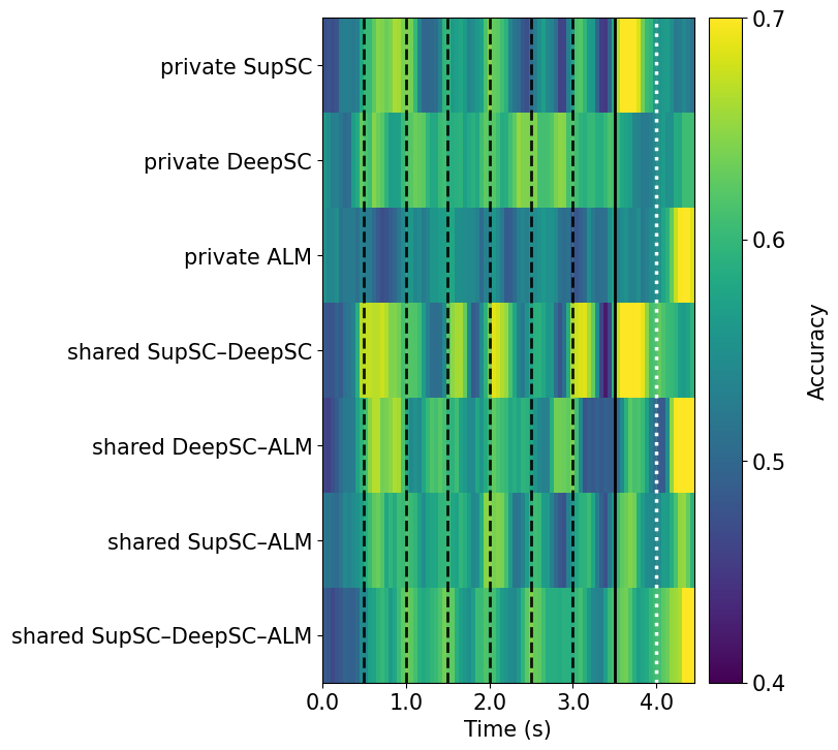}
    \caption{Time-resolved decoding accuracy (mean over 5-fold CV) for each latent subspace. Rows correspond to private and shared subspaces (private SupSC/DeepSC/ALM; shared pairs and all-shared), columns are time. Dashed vertical lines mark cue onsets (every 0.5 s from 0.5–3.0 s), the solid line marks delay start (3.5 s), and the white dotted line marks the start of the decision making period (4.0 s).}
    \label{fig:sc_alm_decoding}
\end{figure}

\begin{figure}
    \centering
    \includegraphics[width=0.75\linewidth]{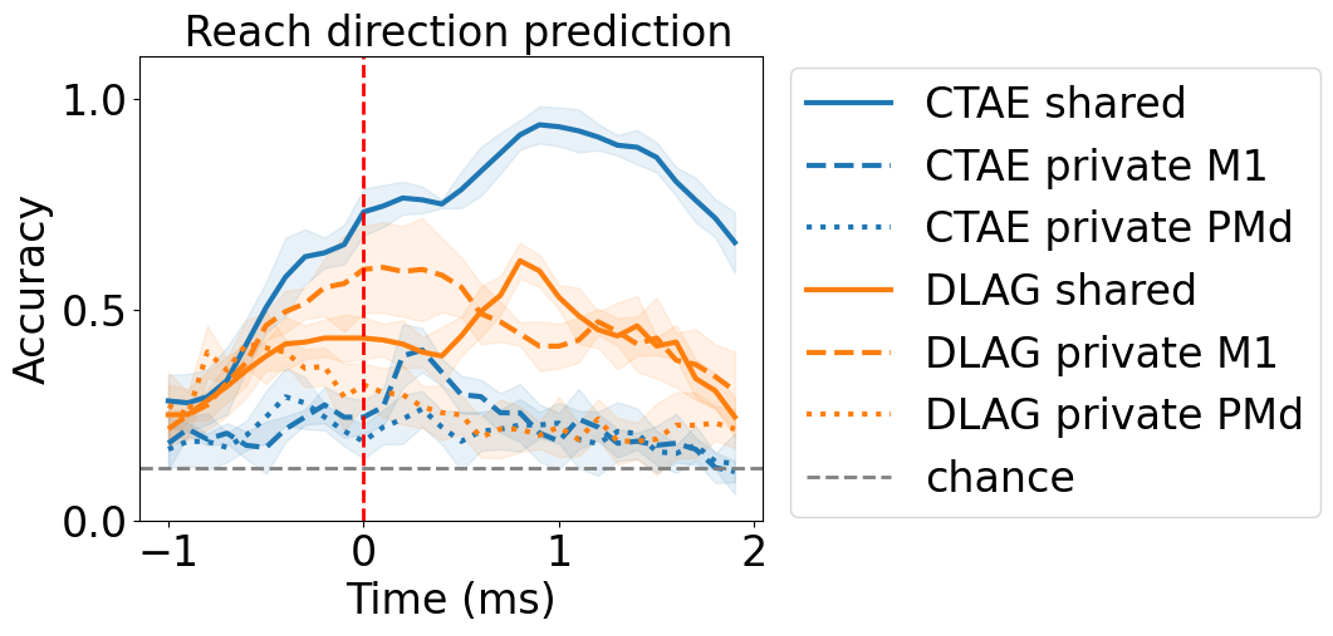}
    \caption{Time-resolved reach-direction decoding accuracy (5-fold CV) for each subspace, across multiple models. Red dashed line represents go-cue. Shared latents from all models show a sharp increase in accuracy following the Go-Cue, with CTAE-shared consistently outperforming others. PMd-specific latents exhibit a pre-Go-Cue rise, indicating anticipatory directional information, whereas M1-specific latents peak post Go-Cue, reflecting direction-specific encoding after movement initiation. Shading represents $\pm 1$ standard error across the different folds.}
    \label{fig:m1pmd_condition_decoding_time}
\end{figure}

\paragraph{Findings: Motor-Circuit dataset.}
Time-resolved reach-direction prediction accuracy shows a pronounced increase following the Go-Cue (Fig. \ref{fig:m1pmd_condition_decoding_time}). This pattern is observed across latents obtained from CTAE, DLAG and DeepCCA. Shared latents from all three models peak in accuracy following the go-cue, indicating presence of higher directional information, with CTAE-shared consistently outperforming those from the other models. In contrast, PMd-specific latents show an  increase in accuracy prior to the go-cue, while, M1-specific latents achieve peak-accuracies post go-cue. These findings suggests that PMd encodes anticipatory directional information, while M1 encodes directional information more strongly during movement. Overall, these results align with previous reports on directional information representation in M1 and PMd  \citep{lara2018conservation}.

\end{document}